%% file: main.tex
\documentclass{article}
\usepackage{arxiv}
\usepackage{booktabs}
\usepackage{multirow}
\usepackage{graphicx}
\usepackage{amsmath}
\usepackage{placeins}
\usepackage{xcolor}
\usepackage{caption}
\usepackage{subcaption}
\graphicspath{{figures/}}


\title{The Identity Trap in EEG Foundation Models:
       A Diagnostic Audit}

\author{%
  Jun-You Lin\thanks{Corresponding author.
    \texttt{linjimmy1003.md10@nycu.edu.tw}.
    ORCID: 0009-0006-3951-3614.} \\
  School of Medicine, National Yang Ming Chiao Tung University, Taipei, Taiwan \\
  \and
  Ying Choon Wu \\
  Swartz Center for Computational Neuroscience,\\
  University of California, San Diego, La Jolla, CA 92037, USA
  \and
  Tzyy-Ping Jung\thanks{ORCID: 0000-0002-8377-2166.} \\
  Swartz Center for Computational Neuroscience,\\
  University of California, San Diego, La Jolla, CA 92037, USA
}

\begin{document}
\maketitle

\input{sections/abstract}
\input{sections/introduction}
\FloatBarrier

\input{sections/related_work}
\FloatBarrier
\input{sections/methods}
\FloatBarrier
\input{sections/results}
\FloatBarrier
\input{sections/discussion}
\FloatBarrier
\input{sections/conclusion}
\FloatBarrier

\input{sections/statements}
\FloatBarrier

\input{sections/appendix}
\FloatBarrier

\bibliography{references}

\end{document}

%% file: sections/abstract.tex
\begin{abstract}
\noindent
\textbf{Objective.}
EEG foundation models (FMs) report strong headline accuracy on
clinical resting-state EEG. However, high accuracy under
subject-disjoint cross-validation remains ambiguous: it can
reflect a genuine clinical biomarker, or subject-identity features
that correlate with the label in this cohort. We name
this ambiguity the \emph{Identity Trap} and ask whether it can be
diagnosed at the representation level before fine-tuning.

\textbf{Approach.}
We propose \textsc{FMScope}, a frozen-representation pre-flight
protocol packaging five diagnostics: variance decomposition,
subject-axis erasure, aperiodic \(1/f\) ablation, layer-wise label
probing, and within-subject direction consistency. We apply it to three pretrained
transformer FMs (LaBraM, CBraMod, REVE) across four public
resting-state datasets (mental arithmetic, sleep deprivation,
Alzheimer's and frontotemporal dementia, trait stress) in an
\emph{a priori} \(2{\times}2\) layout: subject relation of label
\(\times\) presence of a consensus cross-subject EEG marker.

\textbf{Main results.}
(i) The Identity Trap is universal across all three FMs: the frozen
subject-variance fraction is 13--89\(\times\) a random-Gaussian
null in 12 of 12 pairs, rising in all 12 under fine-tuning
(\(+10\) to \(+63\)\,pp). This dominance lies on a removable linear
axis: erasing it significantly improves label decoding where the
label varies within subject (\(+6\) to \(+12\)\,pp in primary cells;
\(+4\) to \(+27\)\,pp across four external consensus-marker cohorts;
one-sided sign test \(p<10^{-3}\)). (ii)
Aperiodic \(1/f\) is one
identifiable subject carrier: removing it drops the subject probe
by \(9\)--\(19\)\,pp uniformly on LaBraM and CBraMod. REVE
saturates subject identity with no measurable
aperiodic dependence and a nonlinearly-decodable residual after
linear erasure: the Identity Trap is universal, but its carrier and
linear removability are model-specific. (iii)
Fine-tuning amplifies label-variance only in cells with a
literature-established cross-subject EEG marker (\(+0.6\) to
\(+8.4\)\,pp). No-consensus cells span zero, indicating no label
signal for fine-tuning to amplify.

\textbf{Significance.}
The Identity Trap is a physically-grounded instance of shortcut
learning: the preferred cue
has a measurable physiological component of the input, not a
pure statistical artifact, and subject-disjoint splitting alone cannot
rule it out. \textsc{FMScope} thus separates gains that reflect a
biological marker from those that reflect subject identity.

\vspace{0.4em}
\end{abstract}

%% file: sections/introduction.tex
\section{Introduction}
\label{sec:intro}

EEG foundation models (FMs) are pretrained on large unlabeled EEG
corpora with self-supervised masked-modeling objectives, and have
been proposed as a general substrate for clinical
electroencephalography \citep{labram2024,cbramod2025,reve2025}. On downstream tasks with
established within-subject neural contrasts (motor imagery,
event-related potentials, sleep staging), reported performance is
strong \citep{eegfmbench2025,adabrain2025,eegbench2025}, building
on a decade of cross-subject classification protocols
\citep{lotte2018}. The picture fragments on small-\(N\) resting-state
EEG (rsEEG), the setting that matters most for psychiatric and
neurodegenerative biomarkers. On
cohorts of comparable size and recording quality, FM performance
varies widely across clinical labels
\citep{brain4fms2026}; what controls this variation has not been
addressed at the representation level. Consider the self-reported chronic-stress dataset of
\citet{komarov2020stress}. A prior FM evaluation on this dataset
reports a peak balanced accuracy of \(0.9047\), using a fixed
80/10/10 train/val/test split in which the same subjects appear
in different folds (best of four data-splitting seeds; the worst
seed reaches \(0.67\)) \citep{wang2025theory}. On the same dataset,
under subject-disjoint cross-validation, we observe \(0.43\)--\(0.50\)
across three FMs and five classical baselines
(Sec.~\ref{sec:results_baseline}). Both numbers can be correct
under their own protocols. Neither tells us what the FM has
actually learned about the label.

A single accuracy number cannot resolve this ambiguity. High
balanced accuracy under subject-disjoint cross-validation is
consistent with at least three readings:
(i) the FM has captured a genuine cross-subject EEG marker of
the clinical label;
(ii) the FM has captured stable physiological subject traits that
happen to co-vary with the label in this cohort; or
(iii) the FM has captured an entanglement of the two that does
not separate at the read-out.
Existing benchmarks enumerate scores
\citep{brain4fms2026,eegfmbench2025,adabrain2025,eegbench2025};
protocol critiques identify subject-identity leakage under
trial-level cross-validation as one inflation source
\citep{brookshire2024leakage}. The underlying tension is not EEG-specific. Resting-state fMRI
fingerprinting shows that stable between-individual differences
in brain activity are sufficient to identify subjects
\citep{finn2015}, and that the connections that identify a subject
and the connections that predict behavior occupy different
functional systems of the connectome \citep{mantwill2022}.
Together, these results suggest a recurring competition, across
brain-imaging modalities, between stable subject-identifying
structure and the task-related signal that biomarkers depend on. Both lines of work establish that the problem exists. Neither tells us, for a
specific cohort \(\times\) FM pair, which of the three readings is
operative at the representation level.

We approach this question through one well-characterized
physiological component of the EEG spectrum: the aperiodic \(1/f\)
background. The standard FOOOF decomposition separates the EEG
power spectrum into two parts: a broadband \(1/f^\chi\) component
(the aperiodic background itself) and narrow periodic peaks at
canonical bands such as theta and alpha
\citep{donoghue2020fooof}. Periodic peaks carry transient,
task-related state information of the kind clinical biomarkers
typically index. The aperiodic background reflects more stable
properties of the recording: cortical excitation--inhibition
balance and vigilance state \citep{gao2017inferring}, and
electrode-level features that vary between individuals and persist
across sessions \citep{kopcanova2023aperiodic}. In ordinary EEG analysis, researchers treat the aperiodic
background as a per-subject nuisance and remove it by fitting a
parametric model before they analyze the periodic peaks. FMs are
pretrained without any subject-aware objective, so it is unclear
whether their representations remove the aperiodic component in
the same way, or instead keep it as an axis that encodes subject
identity. This raises a specific, correlational question: do EEG FM
representations on small-\(N\) rsEEG \emph{co-encode} the aperiodic
\(1/f\) background with subject identity along the same
representational directions? We can test it on LaBraM and CBraMod;
on REVE the test is inconclusive, because REVE differs from the
other two FMs along five design axes at once
(Sec.~\ref{sec:limitations}). Independently of this carrier
question, fine-tuning shows a cell-conditional pattern: it
amplifies subject-related variance in every cell, but amplifies
label-related variance only in cells where the literature has
already established a cross-subject neural marker.

\begin{figure*}[!htbp]
\centering
\includegraphics[width=0.95\linewidth]{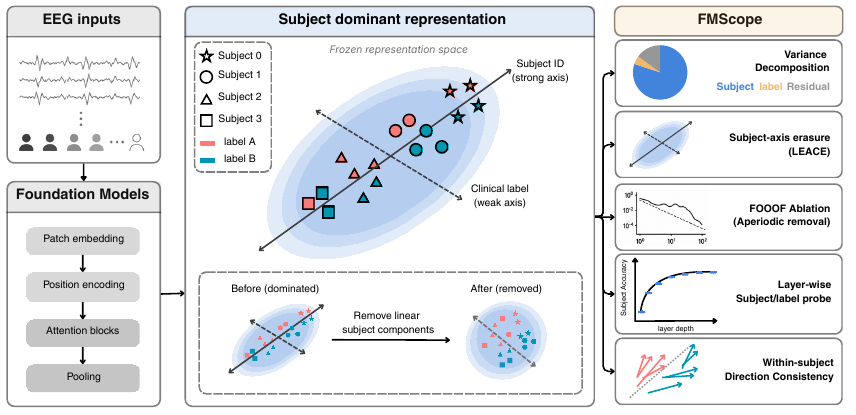}
\caption{\textbf{\textsc{FMScope} overview.} Five frozen-representation
diagnostics applied to embeddings from a pretrained transformer
EEG-FM. Two of the five establish the Identity Trap: variance
decomposition and subject-axis erasure (LEACE). The other three
characterize its origin and
structure: aperiodic input ablation, layer-wise subject/label probe,
and within-subject direction consistency. Center: subject identity
forms the dominant axis of the frozen representation and the clinical
label a weaker one; the inset shows closed-form removal of the linear
subject components (LEACE). Colors and shapes in the embedding feature
space schematically represent variations contributed by cognitive
labels and individual subjects, respectively. Per-tool scope
conditions and details in Sec.~\ref{sec:methods}; results in
Sec.~\ref{sec:results}.}
\label{fig:toolkit_overview}
\end{figure*}

We test this hypothesis on four public small-\(N\) clinical rsEEG
datasets chosen \emph{a priori} to populate a \(2{\times}2\)
sampling layout (subject relation of the label \(\times\) presence
of a consensus cross-subject EEG marker) across three pretrained
transformer FMs (LaBraM, CBraMod, REVE). We package the
diagnostics required for the test as \textsc{FMScope}
(Fig.~\ref{fig:toolkit_overview}), a frozen-representation
framework with explicit scope conditions per tool. We make four
contributions.

\textbf{First}, an empirical finding we term the \emph{Identity
Trap}: across 12 (cell \(\times\) FM) frozen pairs the
subject-variance fraction is 13--89\(\times\) a matched
random-Gaussian null; under fine-tuning, subject-variance
fraction rises in all 12 pairs by \(+10\) to \(+63\) percentage
points (pp). This dominance is confined to a removable linear axis:
closed-form subject-axis erasure drives a linear subject probe to
chance in all 12 pairs, and where the label varies within subject,
erasing identity significantly improves label decoding (\(+6\) to
\(+12\)\,pp in the primary cells; \(+4\) to \(+27\)\,pp across four
external consensus-marker cohorts; one-sided sign test
\(p < 10^{-3}\)).

\textbf{Second}, a representational correlate of the
\(1/f\)--subject co-encoding hypothesis: removing the aperiodic
\(1/f\) component drops a linear subject probe by \(9\) to \(19\)
pp uniformly across all four cells on LaBraM and CBraMod. REVE
shows no measurable aperiodic dependence; the LaBraM-and-CBraMod
group differs from REVE along five concurrent design axes
(Sec.~\ref{sec:limitations}), so we report this two-versus-one
pattern descriptively rather than as a mechanism claim.

\textbf{Third}, a cell-conditional outcome map: fine-tuning
amplifies label-variance only in cells with a consensus
cross-subject EEG marker (Mann--Whitney \(U\), one-sided
\(p = 0.0022\), \(n = 12\)), and the layer-wise label probe
descends monotonically toward chance in the no-consensus trait
cell on all three FMs.

\textbf{Fourth}, the \textsc{FMScope} diagnostic framework
itself, including its per-tool scope conditions, plus the
clinical/protocol guidelines that follow from the three findings
above: recording trait-cell labels as within-subject contrasts
where the state allows, and seeking external physiological
validation where it does not; verifying that within-subject
classifier directions agree across subjects before any BCI
calibration claim; and a frozen-feature
pre-flight that returns a per-cell verdict
(Tab.~\ref{tab:verdict_matrix}) before any fine-tuning compute is
spent.

%% file: sections/related_work.tex
\section{Related Work}
\label{sec:related}

\paragraph{EEG foundation models.}
We anchor on three open-weight backbones.
\textbf{LaBraM} \citep{labram2024} feeds raw EEG patches through a
temporal CNN into a transformer encoder. The pretraining target is
a discrete vector-quantized code for each patch; a separate decoder
maps each code back to the patch's Fourier amplitude and phase, so
the encoder must learn features that support Fourier reconstruction.
It is pretrained on \(\sim\!2{,}500\) hours of mixed EEG.
\textbf{CBraMod} \citep{cbramod2025} uses a criss-cross transformer
that factorizes spatial and temporal attention into two parallel
mechanisms. Its patch embedding adds two branches: a temporal CNN
and an FFT-derived energy vector. Pretraining reconstructs raw EEG
patches under an MSE loss. It is pretrained on a cleaned
\(\sim\!9{,}000\)-hour subset of the Temple University EEG Corpus
(TUEG total: \(\sim\!15{,}000\) subjects, \(\sim\!27{,}000\) hours). \textbf{REVE} \citep{reve2025} is a spatio-temporal transformer
that uses 4D Fourier sinusoidal positional encoding and a linear
patch embedding on the raw signal. Pretraining reconstructs the raw
EEG under an L1 loss, with an additional attention-pooling secondary
task. It is pretrained on \(\sim\!60{,}000\) hours of EEG from 92
datasets covering \(25{,}000\) subjects. All three use masked-modeling self-supervision
but differ along at least five design axes (target representation,
patch embedding, reconstruction loss, positional encoding,
pretraining corpus diversity), and we therefore report cross-FM
contrasts descriptively rather than as mechanism claims. All three report strong
downstream performance on event-related EEG benchmarks
\citep{eegfmbench2025,adabrain2025}, but none has been characterized
at the representation level on small-\(N\) clinical resting-state
cohorts.

\paragraph{Subject leakage and evaluation protocols.}
Prior critiques have established that trial-level cross-validation
in clinical EEG inflates accuracy through subject leakage at the
train--test boundary
\citep{brookshire2024leakage}, and recent
benchmarks
\citep{eegfmbench2025,adabrain2025,eegbench2025,brain4fms2026}
broadly adopt subject-disjoint splitting (with task-specific
exceptions, e.g.\ subject-dependent splits retained for emotion
recognition in EEG-FM-Bench). These works document the
inflation; they do not characterize what the FM has learned in
place of the leaked subject signal once subject-disjoint splitting
is enforced. Our starting point is the ambiguity that remains even after
subject-disjoint splitting: accuracy could still reflect
subject-correlated features that happen to co-vary with the
clinical label in a particular cohort. We ask, at the
representation level, what those features actually are. A concurrent
study \citep{tang2026eegfm} asks the same of EEG-FMs and also uses
LEACE-style erasure as a diagnostic, but applied to a lexicon of
hand-crafted neuro-features (band power, connectivity, entropy).
It reports that removing those features degrades decoding and does not examine
the subject-identity axis. We erase the subject axis instead and find
that its removal can improve consensus-marker decoding, the
complementary direction.

\paragraph{Prior cross-subject EEG markers for clinical labels.}
Prior research documents resting-state or task-related EEG features that have been used as cross-subject biomarkers. For mental
arithmetic, a substantial corpus links frontal-midline theta and
parieto-occipital alpha modulation to cognitive task load (including
mental arithmetic)
\citep{klimesch1999,gevins1997,klimesch2012update}.
For Alzheimer's and frontotemporal dementia, an expert-panel
consensus \citep{babiloni2021expertpanel} cites posterior alpha
peak-frequency and power reduction as a candidate clinical
biomarker; a recent two-cohort replication
\citep{kopcanova2023aperiodic} reports that the AD-versus-HC
spectral signature is ``purely oscillatory'' and that aperiodic
features do not differ between groups. The aperiodic \(1/f\) slope
therefore remains a contested sub-component for AD.
Comparable expert-consensus markers do not exist for our other two
labels: frontal alpha asymmetry as a stress or depression marker
remains contested \citep{reznik2018faa,vandervinne2017faa}, and the
sleep-deprivation dataset paper \citep{xiang2025sleepdep} tabulates
individual differences in sleep quality and traits alongside the
rested-versus-deprived recordings without identifying a
cross-subject EEG marker.
This asymmetry across the four labels is what motivates our
\emph{a priori} sampling layout (Sec.~\ref{sec:methods_2x2}).
None of these works ask whether a pretrained FM's
representation has actually learned the marker; that is the
question we take up in Sec.~\ref{sec:results_aperiodic} via
aperiodic ablation of the input.

\paragraph{Aperiodic \(1/f\) background as a subject feature in EEG.}
The aperiodic \(1/f\) slope and offset are parametrized by FOOOF
\citep{donoghue2020fooof}. \citet{demuru2020eegfingerprint}
demonstrate that these aperiodic features are stable within an
individual across sessions and distinguish subjects independently
of task-evoked oscillations, in parallel to
functional-connectivity fingerprinting on fMRI \citep{finn2015}
and earlier EEG biometric work \citep{campisi2014,marcel2007}.
\citet{demuru2020eegfingerprint} specifically report that
handcrafted aperiodic features identify subjects with higher
accuracy than canonical band-power features, and remain consistent
across eyes-open and eyes-closed conditions. This makes the
aperiodic \(1/f\) one identifiable subject-specific carrier in
classical EEG features. The fMRI fingerprinting work cited above
documents the same between-subject stability for that modality,
without specifying which spectral component carries it. A follow-on connectome analysis further reports that
the edges discriminating individuals show no single-edge overlap with
the edges predicting behavior across cognitive, language, and motor
variables, indicating that subject-identifying and label-relevant
signals can dissociate at the network-feature level
\citep{mantwill2022}. To our knowledge, no prior work has asked which spectral component
carries that subject identity inside a pretrained EEG-FM, nor
whether it survives self-supervised pretraining on cross-subject
corpora.
Sec.~\ref{sec:results_aperiodic} addresses this question via
FOOOF-aperiodic input ablation as a correlational intervention on
the frozen representation.

\paragraph{Mechanism hypotheses for FM representation bias.}
Three lines of prior ML work motivate, but do not establish, the
mechanisms by which a self-supervised FM might preferentially
encode stable physiological subject features over transient
task-related features. \citet{geirhos2020shortcut} document that
neural networks tend to exploit dataset-stable cues that correlate
with labels (\emph{shortcut features}) over invariant
discriminative cues. Under an MSE (squared-error) reconstruction loss, masked-modeling
objectives put most of their loss on the high-variance components
of the input, and the model is widely conjectured to spend
proportionally more capacity on those same components. In EEG, the
broadband aperiodic \(1/f\) baseline carries most of the
low-frequency spectral variance by construction. \citet{reve2025}
explicitly motivate their L1 (rather than MSE) reconstruction loss
as a counter to the L2 sensitivity to noise and outliers in EEG.
This is one concrete design choice on which our three FMs differ.
Linear-network analyses
\citep{saxe2019linear} additionally show that gradient descent on a
linear network preferentially learns the largest-singular-value
directions of the input--output correlation first, and we conjecture
that an analogous direction-ordering operates under small-\(N\)
supervised fine-tuning of the FM head. Each of these is a candidate contributing factor
for our findings; we do not isolate any of them experimentally.

\paragraph{Multi-axis taxonomies in EEG-FM benchmarking.}
Existing aggregate benchmarks organize EEG-FM evaluation either as
a single-dataset case study or as a leaderboard that aggregates
across many tasks and datasets. EEG-FM-Bench
\citep{eegfmbench2025}, AdaBrain-Bench \citep{adabrain2025},
EEG-Bench \citep{eegbench2025}, and Brain4FMs
\citep{brain4fms2026} report dozens of task--dataset cells and
broadly converge on the observation that FMs are not uniformly
superior across clinical cross-subject tasks
\citep{aristimunha2506eeg}: AdaBrain-Bench reports
that on clinical monitoring tasks ``foundation models perform
comparable or even worse than traditional models''
\citep{adabrain2025}. An independent review of ten early
EEG-FMs reaches a parallel conclusion at the methodology level:
that evaluation strategies across the field remain
heterogeneous and limited, and that standardized, scaled
evaluations are a prerequisite for assessing practical
off-the-shelf utility \citep{kuruppu2025critique}. Their taxonomies
are organized along axes orthogonal to ours (task category,
pretraining objective, or fine-tuning strategy), and none ask
which physiological component of the EEG spectrum the learned
representation aligns with. Three FMs and four datasets,
selected \emph{a priori} to span complementary labeling
outcomes, support representation-level diagnostics that read
each pair against the literature-fixed properties the cell
carries.

%% file: sections/methods.tex
\section{Methods}
\label{sec:methods}

This section covers the sampling layout, feature extraction
and evaluation protocol, the fine-tuning recipe, and the five
\textsc{FMScope} diagnostics.

\subsection{The $2\times 2$ sampling layout}
\label{sec:methods_2x2}

We assign the four datasets a priori to a $2\times2$ layout
cross-classifying (a) subject relation of the label and (b) consensus
cross-subject marker. \emph{Axis A} is read from dataset structure:
\textbf{within-subject paired} when each subject contributes recordings
under both classes (EEGMAT, SleepDep) versus \textbf{subject-label
trait} when the label is fixed per subject (ADFTD; Stress under
per-recording DASS-21 binarization \citep{lovibond1995}). For Stress,
14 of 17 subjects carry a single label across all their recordings;
the 3 mixed-cutoff subjects are kept for the headline benchmark and
dropped from any mechanistic diagnostic that requires a subject-level
label. \emph{Axis B} is a literature-anchored a priori expectation. The
\textbf{consensus} column comprises labels for which prior
peer-reviewed EEG work has established a replicable cross-subject
signature: frontal-midline $\theta$ and parieto-occipital $\alpha$
modulation for mental arithmetic (EEGMAT;
\citep{klimesch1999,gevins1997,klimesch2012update}), and posterior
$\alpha$ peak-frequency reduction for AD/FTD (ADFTD;
\citep{babiloni2021expertpanel} expert-panel consensus). The
\textbf{no-consensus} column comprises labels for which no such
signature has been replicated: chronic stress (Stress; frontal alpha
asymmetry contested, \citep{reznik2018faa,vandervinne2017faa}) and
sleep deprivation (SleepDep; the dataset paper reports no candidate
cross-subject EEG marker, \citep{xiang2025sleepdep}). The
$2{\times}2$ layout is fixed before any FM is trained, and we state
its falsifiable consequence in advance: fine-tuning should amplify
label-variance preferentially in consensus cells (tested in
Sec.~\ref{sec:results_regimes}). Tab.~\ref{tab:cell_assignment}
summarizes the assignment.

\input{tables/main/table1_cell_assignment.tex}

\subsection{Per-dataset specifications}
\label{sec:methods_datasets}

All four datasets are small-$N$ ($N \le 65$) resting-state (or
rest-plus-task) EEG at 19--30 channels, $200$\,Hz after per-dataset
resampling. We standardize preprocessing across cells: per-channel
mean subtraction, $1$--$45$\,Hz zero-phase Butterworth band-pass,
$5$\,s non-overlapping epochs.

\textbf{EEGMAT}: 36 subjects $\times$ 72 recordings (3-min
eyes-closed rest vs.\ 1-min serial-subtraction arithmetic);
19-channel 10--20 montage \citep{eegmat}; within-subject paired,
consensus marker.
\textbf{ADFTD}: 65 subjects (AD = 36, HC = 29), one recording per
subject \citep{adftd}; restricted to AD vs.\ HC for Axis B
alignment with the AD-specific prior \citep{babiloni2021expertpanel};
19-channel 10--20 montage; subject-label trait, consensus marker.
\textbf{SleepDep}: the dataset of \citet{xiang2025sleepdep} has 71
participants; 38 of them contributed an eyes-closed recording. We
use that eyes-closed subset and exclude 2 subjects whose session
files are corrupted, leaving $36$ subjects $\times$ $72$ recordings
(baseline vs.\ $24$\,h sleep deprivation); 19-channel;
within-subject paired, no-consensus marker.
\textbf{Stress-DASS}: 17 subjects $\times$ 70 recordings with
per-recording DASS-21 self-report \citep{komarov2020stress};
30-channel; subject-label trait, no-consensus marker.

\subsection{Foundation model feature extraction and per-FM input
normalization} \label{sec:methods_fm}

We evaluate three open-weight EEG foundation models spanning
complementary pretraining objectives and tokenization strategies:
LaBraM \citep{labram2024}, CBraMod \citep{cbramod2025}, and REVE
\citep{reve2025}. Each backbone receives a $5$\,s by $C$-channel
input at 200\,Hz and emits a pooled embedding: a single fixed-length
vector summarizing that 5\,s window, of dimension $d \in \{200, 200,
512\}$ depending on the backbone. The backbone determines the
pooling rule, and we do not re-pool the data. LaBraM and CBraMod
both produce their embedding by mean pooling over encoder output
tokens: LaBraM averages patch tokens after stripping a prepended
classification (CLS) token (matching the release-default head described in
\citealp{labram2024}~§2.1), and CBraMod applies
2D adaptive average pooling over the (channel, patch) grid
(one of the head variants provided in the CBraMod release). REVE returns the attention-pooling
secondary-task token (a learnable query that attends over all
channel--patch tokens, per \citealp{reve2025}~§2.4).

\paragraph{Input normalization.} All three backbones receive raw
microvolt-scale input and each applies its release-default internal
rescaling before its patch embedding. We hold the input contract
fixed across cells and do not sweep it.

\paragraph{Feature extraction modes.} Frozen linear probe (LP)
freezes the backbone (no gradient) and passes the pooled embedding
through a percentile-based clip (fit per dimension on each fold's
training data), a standard scaler, and an $L_2$-penalized logistic
regression with balanced class weights, all held at fixed canonical
values.
Per-window posterior probabilities are
mean-pooled within a recording and thresholded at $0.5$ to produce
the recording-level decision; the threshold is fixed by design and
not tuned. Full
fine-tuning (FT) unfreezes the backbone and adds a single linear
classification head trained end-to-end under the recipe in
Sec.~\ref{sec:methods_ft}. All three backbones expose the same interface. They take a
multi-channel EEG epoch as input and return a pooled embedding, so
the training loop is backbone-agnostic.

\subsection{Evaluation protocol}
\label{sec:methods_eval}

\paragraph{Primary protocol: subject-disjoint 5-fold CV.} All
cells report balanced accuracy (BA), defined as the unweighted mean
of per-class recall, which we use because the cells are
class-imbalanced. Cross-validation is subject-stratified group
$K$-fold with $K = 5$, grouped by subject so that no subject
contributes recordings to both the train and test folds of a given
split \citep{brookshire2024leakage}. We aggregate at the
recording level by averaging per-window posteriors within a
recording, thresholding at $0.5$, and comparing to the recording
label. This is
the single evaluation rule used for the sampling layout axis B
classification and the baseline performance table
(Tab.~\ref{tab:master_performance}).

\paragraph{Multi-seed requirement.} Small-$N$ cells exhibit
non-trivial seed-to-seed FT variability under deterministic cuDNN.
We average every balanced accuracy claim over three FT training
seeds (seeds $\{42, 123, 2024\}$) and report it as mean $\pm$
sample standard deviation; LP additionally averages over a fixed
eight-seed set. Wide confidence intervals reflect the inherent
instability of fine-tuning on small-$N$ cohorts; the relative
performance gaps between tiers remain consistent across seeds.

\subsection{Fine-tuning configuration}
\label{sec:methods_ft}

Each foundation model is fine-tuned under the configuration
published in its original repository, applied uniformly across the
four cells; we do not run a per-dataset hyperparameter sweep. All three FMs share the AdamW optimizer, the cross-entropy loss with
label smoothing, early stopping on a moving-average of held-out
balanced accuracy, and a single linear classification head with two
output units. LaBraM additionally uses the small-scale head
initialization prescribed in its release \citep{labram2024}.

\subsection{Reference baselines (classical and non-FM-deep)}
\label{sec:methods_baselines}

Two reference baselines situate the FM tier in
Tab.~\ref{tab:master_performance}.

\paragraph{Classical handcrafted-feature baseline.} A per-recording
feature vector is built from the time-averaged power spectrum
(Welch, $n_{\text{perseg}}=\min(256, T)$): for each channel, mean
band-power in $\theta$ ($4$--$8$\,Hz), $\alpha$ ($8$--$13$\,Hz)
and $\beta$ ($13$--$30$\,Hz); per-channel $\theta/\alpha$ and
$\theta/\beta$ ratios; and frontal/parietal alpha-asymmetry
$\log P_\alpha^{\text{right}} - \log P_\alpha^{\text{left}}$ over six
electrode pairs (Fp1/Fp2, F3/F4, F7/F8, C3/C4, P3/P4, O1/O2). At
$19$ channels this yields $5\times 19 + 6 = 101$ features. We fit
$L_2$-regularized logistic regression with feature standardization,
inverse regularization strength $C = 1$, and balanced class weights,
under the same subject-disjoint 5-fold CV, three seeds.

\paragraph{Non-FM-deep baselines.} Four convolutional / hybrid
architectures trained from scratch on raw EEG: EEGNet
\citep{lawhern2018}, ShallowConvNet and DeepConvNet
\citep{schirrmeister2017}, and EEG-Conformer
\citep{song2023eegconformer}. Each is trained per dataset under the
same CV splits and seeds, with $z$-score input normalization
(matching the early-BatchNorm convention used in standard
implementations of these architectures). No
foundation-model pretrained weights are used.

\subsection{Diagnostic method specifications}
\label{sec:methods_toolkit}

\begin{figure*}[!htbp]
\centering
\includegraphics[width=0.95\linewidth]{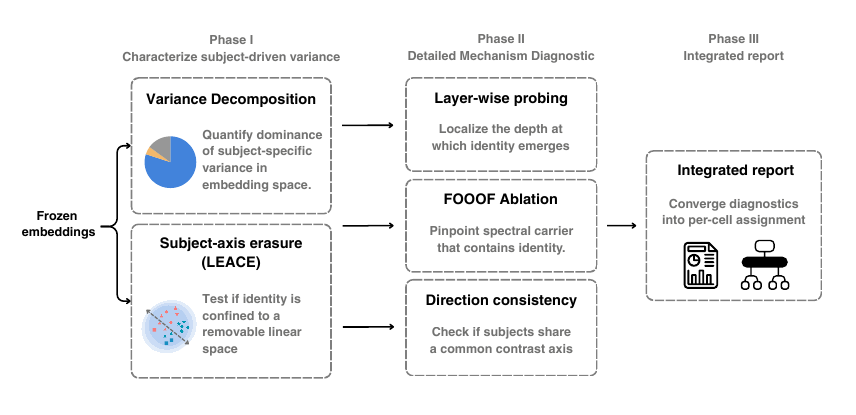}
\caption{\textbf{\textsc{FMScope} diagnostic pipeline.} The framework evaluates frozen representations from EEG foundation models across three sequential phases. \textbf{Phase I} establishes the existence of the Identity Trap by quantifying subject-variance dominance and testing its linear removability via least-squares concept erasure (LEACE). \textbf{Phase II} characterizes the underlying mechanisms using three tools: localizing the depth of subject-identity emergence (layer-wise probing), isolating the physiological carrier via spectral input ablation (FOOOF), and evaluating task-axis alignment across subjects (direction consistency). \textbf{Phase III} synthesizes these diagnostic outputs into an integrated report, converging on a final cell-level verdict for the specific cohort-model pair.}
\label{fig:pipeline}
\end{figure*}

\textsc{FMScope} comprises five frozen-representation diagnostics,
organized by the question each answers. Two of these establish the Identity
Trap itself: whether subject identity dominates the representation
(variance decomposition, Sec.~\ref{sec:methods_variance}) and
whether that dominance is confined to a linearly removable axis
(subject-axis erasure, Sec.~\ref{sec:methods_leace}). The other three
characterize its origin and structure: its spectral carrier
(aperiodic ablation, Sec.~\ref{sec:methods_anchor}), the depth at
which it becomes linearly separable (layer-wise probe,
Sec.~\ref{sec:methods_layerwise}), and whether subjects encode the
task contrast along a shared direction (direction consistency,
Sec.~\ref{sec:methods_wsci}). Each diagnostic carries a scope
condition stating the cells in which it returns a defined answer
and is read only where that scope is met, so a cohort need not
exercise all five.

\subsubsection{Variance decomposition of frozen representations}
\label{sec:methods_variance}

For each (dataset, FM) we decompose per-window embedding variance
via a crossed two-factor sum-of-squares partition with label
and subject as factors; this is the marginalization step of
demixed PCA \citep{kobak2016dpca}.
\begin{equation}
\mathrm{SS}_{\text{total}} \;\approx\;
\mathrm{SS}_{\text{label}} +
\mathrm{SS}_{\text{subject}} +
\mathrm{SS}_{\text{residual}},
\label{eq:ss_decomp}
\end{equation}
where, for a factor $g \in \{\text{label}, \text{subject}\}$ with
groups indexed by $c$, we compute the between-group SS marginally
over the other factor as
\begin{equation}
\mathrm{SS}_{g} \;=\;
\sum_{c} n_{c}\,\bigl\lVert
\bar{\mathbf{f}}_{c} - \bar{\mathbf{f}}
\bigr\rVert^{2},
\label{eq:ss_factor}
\end{equation}
with $\bar{\mathbf{f}}_{c}$ the per-group mean embedding,
$\bar{\mathbf{f}}$ the grand mean, $n_c$ the group window count, and
$\mathrm{SS}_{\text{residual}}$ obtained by subtraction and clipped at
zero. We operate at the window level so the partition is defined for
single-session trait cells (ADFTD: each recording yields $\geq 2$
windows). Cluster-bootstrap CIs over subjects ($B = 2{,}000$)
preserve the within-subject correlation structure
\citep{field2007clusterbootstrap}. Reading the partition is cell-conditional. In within-subject paired
cells the two factors are orthogonal so the fractions sum to
$<\!1$; in trait cells the subject factor structurally contains
the label factor, which naturally allows the marginal fractions to
sum to $>\!1$ because they represent overlapping variance
components. We therefore emphasize
the within-cell label-fraction shift (FT $-$ frozen) rather than
absolute fractions. Let
$f_{\mathrm{label}} = \mathrm{SS}_{\text{label}} /
\mathrm{SS}_{\text{total}}$ and
$f_{\mathrm{subj}} = \mathrm{SS}_{\text{subject}} /
\mathrm{SS}_{\text{total}}$ denote the label- and subject-variance
fractions; throughout, $\Delta$ denotes the change after the
relevant intervention (here fine-tuning; in
Sec.~\ref{sec:methods_anchor}, aperiodic removal), so
$\Delta f_{\mathrm{label}}$ is the FT $-$ frozen change in
label-variance fraction and $\Delta f_{\mathrm{subj}}$ is the
analogous change in subject-variance fraction. Scope condition:
all cells; for Stress the 3/17 mixed-cutoff subjects are dropped,
leaving a 14-subject strict-trait subset.

\paragraph{Column-effect test on $\Delta f_{\mathrm{label}}$.}
Across the 12 (cell, FM) pairs we contrast the 6 consensus-marker
pairs against the 6 no-consensus pairs on
$\Delta f_{\mathrm{label}}$ using a one-sided Mann--Whitney U
test; the test is non-parametric and distribution-free,
appropriate for the small sample. The 12 pairs share data within
cell (cell-level $n = 4$ with 3 FMs nested per cell), so we read
U as an effect-size summary (we report the rank-biserial $r$ in
the Results) rather than as axis-level inference.

\subsubsection{Subject-axis erasure}
\label{sec:methods_leace}

Variance decomposition measures how much of the representation
subject identity occupies; erasure tests whether that identity is
confined to a removable linear axis and at what cost to the
clinical label. We apply LEACE \citep{belrose2023leace}, least-squares concept
erasure: the minimum-displacement affine map that renders a target
concept linearly unpredictable. For per-window embeddings $\mathbf{X}$ and
one-hot subject labels $\mathbf{Z}$, LEACE removes the subspace
spanned by $\mathbf{W}\Sigma_{\mathbf{XZ}}$, with
$\mathbf{W} = \Sigma_{\mathbf{XX}}^{-1/2}$ the whitening transform,
which we estimate under Ledoit--Wolf shrinkage for numerical
conditioning. Centred one-hot labels for $k$ subjects span $k-1$
dimensions, so the erased subspace has rank $k-1$: it is the
between-subject-mean subspace. After erasure no linear classifier
recovers subject identity above chance; the guarantee is linear
only, so we report the identity still recoverable by a nonlinear
probe (a one-hidden-layer MLP) alongside it. The map is undefined
when $k-1 \geq d$, where the subject subspace fills the feature
space; we test this condition and skip the cell when it holds. To
quantify the cost of erasure to the task, we re-run the
recording-level label probe (Sec.~\ref{sec:methods_anchor}) on the
erased features and report $\Delta_{\mathrm{erase}}$, the change in
label balanced accuracy. Scope condition: all cells for the
erasure and nonlinear-residual read-out. $\Delta_{\mathrm{erase}}$
is interpretable only in within-subject paired cells, where the
label varies within subject and is therefore separable from the
erased between-subject subspace; in trait cells the label is a
per-subject constant whose class-mean direction lies inside the
subject subspace, so erasure removes it by construction and
$\Delta_{\mathrm{erase}}$ is undefined. It is read only where the
un-erased baseline also clears chance.

\subsubsection{Spectral Anchor Ablation (FOOOF)}
\label{sec:methods_anchor}

We intervene on the EEG via FOOOF ablation, which modifies the
spectral shape (aperiodic background vs.\ periodic peaks) while
preserving frequency support, and observe how the FM's frozen
probe responds.

For each recording we fit
a per-channel FOOOF decomposition \citep{donoghue2020fooof} on the
$1$--$45$\,Hz band, parameterizing the log-power spectrum as
\begin{equation}
\log_{10}\!\mathrm{PSD}(f) =
\underbrace{b - \chi\,\log_{10} f}_{\text{aperiodic } L(f)}
+ \sum_{n=1}^{N} G_n(f),
\label{eq:fooof_psd}
\end{equation}
where $b$ is an offset, $\chi$ (Greek chi) the aperiodic exponent
(slope of $1/f$ on log-log axes; higher $\chi$ means relatively
more low-frequency power), and each $G_n(f)$ a Gaussian peak in
log-power. With $X(f)$ the per-channel FFT,
$\hat A(f) = 10^{L(f)/2}$ the fitted aperiodic amplitude envelope,
and $P_n(f) = \hat A(f)^2(10^{G_n(f)} - 1)$ the linear-power
contribution of peak $n$, we construct three phase-preserving
reconstructions
\begin{equation}
|\widetilde X(f)| =
\begin{cases}
|X(f)|/\hat A(f) & \text{aperiodic-removed,} \\[1pt]
\sqrt{\max(|X(f)|^2 - \sum_n P_n(f),\,0)} & \text{periodic-removed,} \\[1pt]
|\widetilde X_{\text{per-rem}}(f)|/\hat A(f) & \text{both-removed,}
\end{cases}
\label{eq:fooof_recon}
\end{equation}
each combined with the original phase spectrum and inverted via
FFT.
Each reconstruction is re-extracted through the FM and probed with
two diagnostics. The label probe is a binary $L_2$-penalized
logistic regression on per-window features under subject-stratified
group $K$-fold ($K = 5$). The pipeline matches the LP recipe
(percentile clip + standard scaler) and aggregates to the
recording level by mean-pooling per-window posteriors; balanced
accuracy is averaged over $8$ seeds (the LP seed set). The subject-identity probe is a multi-class classifier
of subject identity (one class per subject) under a 5-fold
\emph{temporal-block} protocol: per subject, the windows are concatenated
across recordings in canonical order and split into 5 contiguous
blocks; fold $f$ tests block $f$ from every subject and trains on the
remaining four blocks. The
classifier is linear discriminant analysis with Ledoit--Wolf
shrinkage of the within-class covariance toward a scaled identity
matrix. The shrinkage stabilizes the covariance estimate when the
number of windows per subject is small. The classifier uses a closed-form
solver with no stochastic optimization, so we run it at a single
seed. Per-fold balanced accuracy is averaged across the 5 folds. On multi-recording cells (EEGMAT, SleepDep, Stress) the blocks
cross recording boundaries, so the probe also tests whether the
subject signal is stable across sessions. ADFTD has only one
recording per subject by design, so its blocks are always within a
single recording.

\paragraph{FT extension.} To extend the intervention from the
frozen representation to the fine-tuned representation, we
re-extract and fine-tune each backbone end-to-end on the
ablated input under the same recipe as
Sec.~\ref{sec:methods_ft} and run both probes (state and
subject) on the resulting test-fold per-window features
concatenated across the 5 CV folds. Per (cell, FM, condition) we
average \(\Delta\)probe\,BA = ablated minus original input over
3 fine-tuning seeds; the FT extension covers both conditions
(aperiodic-removed and periodic-removed).

Output. The diagnostic returns a per-cell $1/f$-role reading
derived from the joint change in label-probe BA and subject-probe BA
under aperiodic removal, where each $\Delta$ is computed as
aperiodic-removed minus original.

Scope condition: all four cells for the label probe and subject
probe at frozen and FT representations.

\subsubsection{Layer-wise probe}
\label{sec:methods_layerwise}

To localize the depth at which subject identity emerges within the
encoder, we replay each FM's forward pass with eight intermediate-depth
captures and apply the same two read-outs used elsewhere in the paper
(temporal-block subject-ID probe; canonical recording-level linear
probe with three CV seeds) at every captured depth. Pooling at each
captured depth uses the backbone's own canonical scheme, so the
final-depth row reproduces the main-paper frozen feature to numerical
precision.

Scope condition: all four cells for both probes.

\subsubsection{Within-subject direction and signal-to-noise
characterization}
\label{sec:methods_wsci}

For within-subject paired cells we measure (i) whether subjects
encode the state contrast along a consistent direction in FM
feature space, and (ii) whether per-subject contrast magnitudes
separate from cross-subject heterogeneity. For each subject $s$ we
form the per-subject contrast vector
\begin{equation}
\boldsymbol{\Delta}_{s} \;=\;
\bar{\mathbf{f}}_{s,1} - \bar{\mathbf{f}}_{s,0},
\label{eq:wsci_delta}
\end{equation}
where $\bar{\mathbf{f}}_{s,y}$ is the window-mean embedding for
subject $s$ under label class $y$. The within-subject direction
consistency index $\bar c$ (WSCI) is the mean pairwise cosine
similarity across the $n$ per-subject contrast vectors,
\begin{equation}
\bar c \;=\;
\binom{n}{2}^{-1} \sum_{i<j}
\frac{\boldsymbol{\Delta}_{i}^{\top}\boldsymbol{\Delta}_{j}}
     {\lVert\boldsymbol{\Delta}_{i}\rVert\,
      \lVert\boldsymbol{\Delta}_{j}\rVert};
\label{eq:wsci}
\end{equation}
high $\bar c$ implies a shared cross-subject axis, $\bar c \approx 0$
implies idiosyncratic directions. The per-subject SNR is a signal-to-noise ratio of mean contrast
magnitude $\sigma_s = \lVert\boldsymbol{\Delta}_s\rVert$ to its
cross-subject standard deviation,
\begin{equation}
\mathrm{SNR}_{\text{per-subj}} \;=\;
\frac{\overline{\sigma_s}}
     {\mathrm{SD}_s\!\left(\sigma_s\right)}.
\label{eq:snr_persubj}
\end{equation}
$\bar c$ and SNR together characterize direction agreement and
magnitude-vs-noise; we do not compose them into a single assessment.
Scope condition: within-subject paired cells only (EEGMAT,
SleepDep); marked inapplicable in trait cells.

\paragraph{Within-cell label-structure detector (cosine PERMANOVA).}
$\bar c$ is undefined in trait cells (no within-subject contrast)
and uninterpretable when subjects do not share a direction
($\bar c \approx 0$). To detect whether a label-associated
signature exists in feature geometry, independent of whether
subjects share a contrast direction, we run PERMANOVA
\citep{anderson2001permanova} on the cosine dissimilarity matrix
of per-window embeddings. The design factor is the labeling unit
(subject for ADFTD; recording within pair block for EEGMAT and
SleepDep; recording for Stress). We use \(999\) permutations per
(cell, FM, state), giving a floor of \(p \approx 0.001\). We read
PERMANOVA as a within-cell test for whether label-related
geometric structure exists at all, not as the axis-level
consensus-vs-no-consensus contrast (which is reported on
$\Delta f_{\mathrm{label}}$ in Sec.~\ref{sec:methods_variance}). Scope condition: all four
cells. Per-cell assessments are tabulated in
App.~\ref{app:permanova}.

%% file: tables/main/table1_cell_assignment.tex
% Tab 1 — 2x2 sampling-layout grid (body only)
\begin{table}[htbp]
\centering
\footnotesize
\caption{\textbf{Dataset assignment to the $2\times2$ sampling layout.}
Rows: subject relation of the label (Axis A, read from dataset
structure). Columns: consensus cross-subject marker (Axis B, fixed a
priori from peer-reviewed EEG literature under the criterion in
Sec.~\ref{sec:methods_2x2}). Each cell gives the dataset, its
literature anchor, and recordings\,/\,subjects.}
\label{tab:cell_assignment}
\begin{tabular}{@{}l l l@{}}
\toprule
 & \textbf{Consensus marker} & \textbf{No-consensus marker} \\
\midrule
\textbf{Within-subject} & \textbf{EEGMAT} & \textbf{SleepDep} \\
\textbf{paired}
 & FMT $\theta$ + occipital $\alpha$
 & no replicated marker \\
 & \citep{klimesch1999,gevins1997}
 & \citep{xiang2025sleepdep} \\
 & 72 rec / 36 subj & 72 / 36 \\
\midrule
\textbf{Subject-label} & \textbf{ADFTD} & \textbf{Stress (DASS)} \\
\textbf{trait}
 & posterior $\alpha$ peak / power
 & FAA contested \\
 & \citep{babiloni2021expertpanel}
 & \citep{reznik2018faa,vandervinne2017faa} \\
 & 65 / 65 & 70 / 17 \\
\bottomrule
\end{tabular}
\end{table}

%% file: sections/results.tex
\section{Results}
\label{sec:results}

The Results proceed in four steps. We start with the baseline
performance picture: full fine-tuning rarely improves over a
frozen linear probe (Sec.~\ref{sec:results_baseline}). To see
what fine-tuning starts from, we then characterize the frozen
representation itself; subject identity dominates, and the
encoder amplifies it into a linearly separable direction within
its first few transformer blocks
(Sec.~\ref{sec:results_subject_dominant}). A natural next
question is which input feature carries that subject identity.
The aperiodic \(1/f\) background is one such carrier on LaBraM
and CBraMod (Sec.~\ref{sec:results_aperiodic}).
Sec.~\ref{sec:results_regimes} summarizes the per-cell evidence
for the final assessment.

\subsection{Baseline performance and the fine-tuning paradox}
\label{sec:results_baseline}

Tab.~\ref{tab:master_performance} reports subject-disjoint balanced
accuracy across the four cells under three FMs (linear probe and
full fine-tuning), four convolutional / hybrid baselines trained
from scratch, and a classical handcrafted-feature LogReg.

\input{tables/main/table2_master_performance.tex}

Fine-tuning rarely beats the frozen linear probe: 10 of 12 pairs
fall within \(\pm 1\)\,pp of the corresponding LP. ADFTD is the
one cell where FM pretraining yields a measurable advantage over
the classical and non-FM-deep baselines. On EEGMAT the classical
RF beats every FM tier. On Stress and SleepDep all four tiers
fall within a \(0.43\)--\(0.57\) band. The Wang \emph{et al.}
\(0.9047\) Stress headline \citep{wang2025theory} does not
reproduce under subject-disjoint cross-validation; the gap is
\(\sim 45\)\,pp.

\subsection{Subject-dominant baseline geometry}
\label{sec:results_subject_dominant}
\label{sec:results_variance}

\begin{figure*}[!htbp]
\centering
\includegraphics[width=\linewidth]{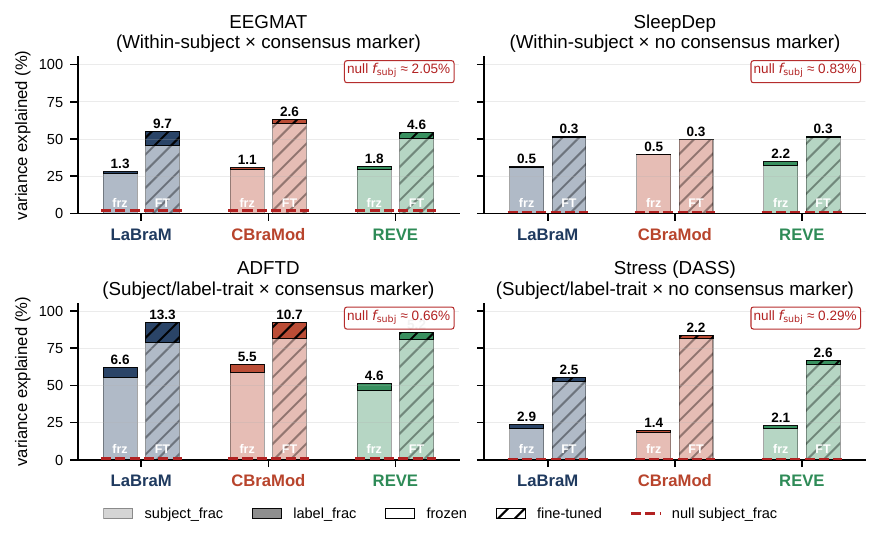}
\caption{\textbf{Variance decomposition across the four cells.}
Window-level subject and label fractions for frozen and fine-tuned
features. Stacked bars: subject (lower) + label (upper); gap to
100\% is residual. Dashed red line marks the matched random-Gaussian
null \(f_{\mathrm{subj}}\) (mean over 20 seeds; per-cell numeric
callout in each panel). Frozen subject fraction exceeds the null by
13--89\(\times\) across 12 (cell, FM) pairs; under fine-tuning,
subject fraction rises in 12 of 12 pairs (largest on Stress:
\(+32\) / \(+63\) / \(+43\)\,pp on LaBraM / CBraMod / REVE), while
label fraction rises only in cells with a consensus cross-subject
marker (per-pair values in Tab.~\ref{tab:variance_full};
null-control values in Tab.~\ref{tab:null_control}).}
\label{fig:variance_2x2}
\end{figure*}

We decompose per-window frozen and fine-tuned features into label,
subject, and residual variance components
(Sec.~\ref{sec:methods_variance}; Fig.~\ref{fig:variance_2x2}).
The frozen subject-variance fraction is 13--89\(\times\) a matched
random-Gaussian null in 12 of 12 (cell $\times$ FM) pairs (per-pair
values in Tab.~\ref{tab:null_control}); fine-tuning increases the
subject fraction in all 12 pairs (\(+10\) to \(+63\)\,pp, largest
on Stress), while \(\Delta f_{\mathrm{label}}\) is positive in 6 of 6
consensus-marker pairs (\(+0.6\) to \(+8.4\)\,pp) and spans zero
in 6 of 6 no-consensus pairs (\(-1.9\) to \(+0.8\)\,pp). The two
\(\Delta f_{\mathrm{label}}\) ranges do not overlap (one-sided
Mann--Whitney U on the 12 values: \(p = 0.0022\), rank-biserial
\(r = +0.94\); Sec.~\ref{sec:methods_variance}).

This subject dominance is a removable linear axis, not diffuse
structure (Tab.~\ref{tab:leace}). Closed-form erasure
(Sec.~\ref{sec:methods_leace}) drives the linear subject probe to
chance in all 12 pairs. A nonlinear probe still recovers part of
the identity, substantial for REVE, which we report alongside.
Whether removing the axis helps the task can be asked only in
within-subject paired cells, where the label varies within
subject. On EEGMAT, the one such cell with an above-chance
baseline, the frozen probes start below the classical baseline
(\(0.847\)); erasure lifts all three FMs and closes this gap, with
LaBraM reaching \(0.875\). On SleepDep (no-consensus, near-chance
baseline) the change is small and inconsistent across FMs, as
expected when little label signal is present. This is a single
clean demonstration; App.~\ref{app:erasure_gen} extends it to five
further cohorts with a within-subject contrast, where
\(\Delta_{\mathrm{erase}}\) stays positive on all three
consensus-marker cohorts and turns mixed-to-negative on the two
without an established marker (positive in 12 of 12 consensus
cell\(\times\)FM values including the primary cell; one-sided sign
test \(p = 2.4\times10^{-4}\)).

\begin{table}[!htbp]
\centering
\small
\setlength{\tabcolsep}{4pt}
\caption{\textbf{Subject-axis erasure on frozen features.}
BA \(\times 100\); chance \(=100/k\); erased subspace rank
\(=k-1\). Subject BA is a linear subject classifier before and
after erasure; nonlinear residual is an MLP after erasure (LEACE
guarantees linear erasure only). \(\Delta_{\mathrm{erase}}\) is the
change in recording-level label BA (3 seeds, mean \(\pm\) SD),
defined only in within-subject paired cells; trait cells (ADFTD,
Stress) are marked ``---'' because the label is a fixed subject
attribute lying inside the subject subspace.}
\label{tab:leace}
\begin{tabular}{llrrrr}
\toprule
\textbf{Cell} & \textbf{FM} & \textbf{Subj.\ pre} & \textbf{Subj.\ post} & \textbf{Nonlin.} & \(\Delta_{\mathrm{erase}}\) \\
\midrule
\multirow{3}{*}{EEGMAT}   & LaBraM  & 79.6 & 0.0 & 1.6  & \(+12.0 \pm 7.4\) \\
                           & CBraMod & 70.5 & 0.0 & 2.5  & \(+6.0 \pm 3.3\)  \\
                           & REVE    & 99.5 & 0.0 & 3.3  & \(+7.4 \pm 0.7\)  \\
\midrule
\multirow{3}{*}{ADFTD}    & LaBraM  & 93.4 & 0.3 & 1.5  & --- \\
                           & CBraMod & 90.6 & 0.3 & 1.4  & --- \\
                           & REVE    & 99.4 & 0.6 & 11.5 & --- \\
\midrule
\multirow{3}{*}{SleepDep} & LaBraM  & 73.2 & 0.1 & 4.3  & \(+1.9 \pm 0.7\) \\
                           & CBraMod & 73.2 & 0.4 & 5.5  & \(+0.9 \pm 4.7\) \\
                           & REVE    & 97.3 & 1.1 & 56.4 & \(+7.4 \pm 2.9\) \\
\midrule
\multirow{3}{*}{Stress}   & LaBraM  & 64.7 & 4.2 & 6.9  & --- \\
                           & CBraMod & 59.0 & 4.0 & 5.2  & --- \\
                           & REVE    & 98.0 & 5.3 & 40.0 & --- \\
\bottomrule
\end{tabular}
\end{table}

\begin{figure*}[!htbp]
\centering
\includegraphics[width=\linewidth]{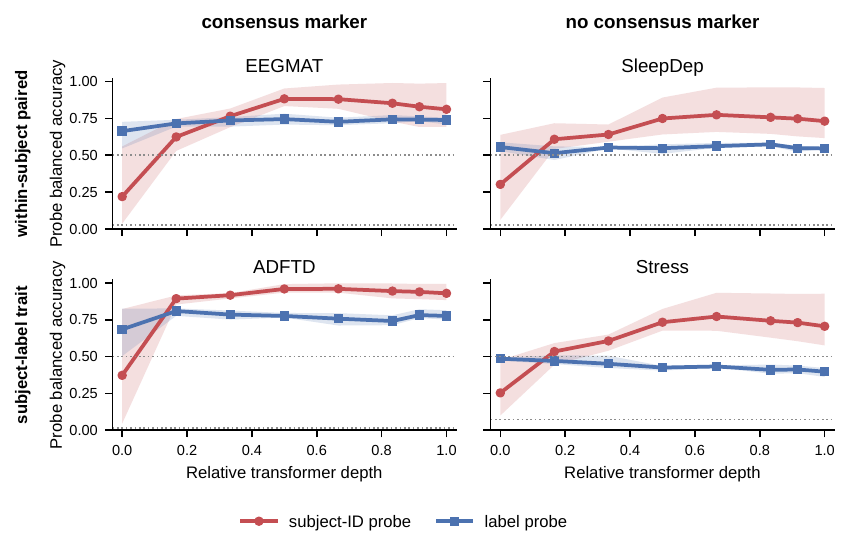}
\caption{\textbf{Layer-wise subject and label probes.} Rows show
the subject relation of the label (within-subject paired on top,
trait on the bottom); columns show the consensus axis (consensus
on the left, no-consensus on the right). Each panel shows two
probes: the temporal-block subject-ID probe (red) and the
canonical recording-level label probe (blue). Lines are the mean
across the three FMs (LaBraM, CBraMod, REVE); shaded bands span
their min--max range (\(n=3\) FMs, an envelope rather than a
confidence interval). The horizontal axis is relative transformer
depth (\(0\) is post-embedding pre-block; \(1\) is the final
pooled feature). Across all four cells the subject probe rises
toward \(0.55\)--\(0.99\) within two to three transformer blocks;
the wide early-depth band reflects FMs that reach this ceiling at
different depths. The label probe is cell-conditional. Dotted
lines mark label chance (\(0.5\)) and the per-cell subject-ID
chance.}
\label{fig:layerwise}
\end{figure*}

A layer-wise re-probe (Sec.~\ref{sec:methods_layerwise};
Fig.~\ref{fig:layerwise}) shows that the subject axis becomes
linearly decodable inside the encoder rather than at the embedding
output: the encoder actively amplifies subject identity into a
separable direction within its first two to three blocks. The
label probe is cell-conditional; the per-cell trajectories feed
the per-cell assessment in Sec.~\ref{sec:results_regimes}.

\subsection{Aperiodic \(1/f\) as a subject-identity carrier}
\label{sec:results_aperiodic}
\label{sec:results_anchor}

\begin{figure*}[!htbp]
\centering
\begin{subfigure}[t]{0.95\linewidth}
  \centering
  \includegraphics[width=\linewidth]{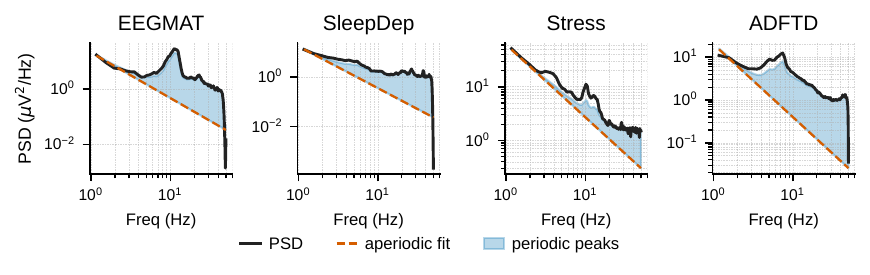}
  \caption{}
  \label{fig:anchor:psd}
\end{subfigure}\\[6pt]
\begin{subfigure}[t]{0.95\linewidth}
  \centering
  \includegraphics[width=\linewidth]{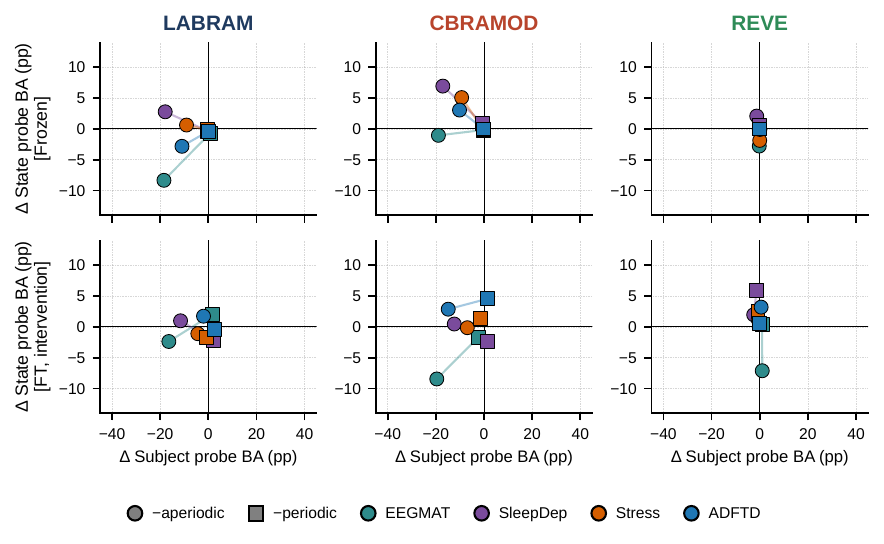}
  \caption{}
  \label{fig:anchor:scatter}
\end{subfigure}
\caption{\textbf{Aperiodic and periodic ablation of the input,
frozen and intervention-FT.}
\textbf{(\subref{fig:anchor:psd})} Representative log-log power
spectrum per cell with FOOOF decomposition: black solid, measured
PSD; orange dashed, \(1/f\) aperiodic fit; blue shading, periodic
peaks (PSD minus aperiodic). The aperiodic fit defines the input
ablation (Sec.~\ref{sec:methods_anchor}).
\textbf{(\subref{fig:anchor:scatter})} \(\Delta\) probe BA under
FOOOF ablation; circles mark \(-\)aperiodic, squares mark
\(-\)periodic, a line within each FM connects the two conditions
per cell, colour codes the cell; top row frozen, bottom row FT
(intervention, 3-seed mean).
Frozen: FOOOF
\(-\)aperiodic on LaBraM and CBraMod uniformly drops the subject
probe by \(9\) to \(19\)\,pp across all four cells; \(-\)periodic
shifts probe BA by \(\leq 0.8\)\,pp on every (FM, cell)
combination; REVE's already-saturated subject probe
(\(\geq 0.93\) baseline) shows no measurable aperiodic dependence.
FT (intervention): re-extracting and
fine-tuning under aperiodic-removed input attenuates the FT
subject probe on 7 of 8 (cell, FM) pairs for LaBraM/CBraMod
(\(-4\) to \(-20\)\,pp); LaBraM \(\times\) ADFTD is borderline
(\(-2.0 \pm 4.7\)\,pp). \(-\)Periodic FT shifts the subject probe
by \(|\Delta| \le 2.5\)\,pp on every (cell, LaBraM/CBraMod) pair,
mirroring the frozen null and locating the FT effect in the
\(1/f\) component rather than narrowband peaks. REVE's FT
bottom-row shifts \(|\Delta| \le 2.5\)\,pp on every cell under
both ablations, preserving the negative control downstream.
Cell-conditional label-probe response is interpreted in the
main text.}
\label{fig:anchor}
\end{figure*}

We intervene on the input EEG via FOOOF spectral-shape ablation
(aperiodic background versus periodic peaks; frequency support
preserved) and observe the FM probe response
(Sec.~\ref{sec:methods_anchor}; Fig.~\ref{fig:anchor}).

\textbf{Removing the aperiodic \(1/f\) component drops the subject
probe uniformly on LaBraM and CBraMod.} On LaBraM and
CBraMod, FOOOF \(-\)aperiodic drops the linear subject probe BA by
\(9\) to \(19\)\,pp across all four cells (per-(cell, FM) values
in Tab.~\ref{tab:fooof_full}; LaBraM:
\(-18.4\) / \(-10.9\) / \(-17.9\) / \(-9.1\)\,pp on EEGMAT / ADFTD
/ SleepDep / Stress; CBraMod: \(-18.9\) / \(-10.2\) / \(-17.1\) /
\(-9.3\)\,pp). The drop is large, uniform across cells of very
different paradigms, and goes in the same direction on both FMs.

REVE's baseline subject probe is already at \(\geq 0.93\) BA and
shifts by \(\leq 1.2\)\,pp under \(-\)aperiodic; we therefore
limit the aperiodic-anchor claim to LaBraM and CBraMod
(Sec.~\ref{sec:limitations}). FOOOF \(-\)periodic shows no effect
across all 12 (cell, FM) pairs (\(\leq 0.8\)\,pp on both probes).
Because both ablations pass through the same FOOOF
decompose-and-invert reconstruction (Eq.~\ref{eq:fooof_recon}) and
differ only in which component is removed, this \(-\)periodic null
rules out the reconstruction round-trip itself as the cause and
attributes the \(-\)aperiodic drop to the removed aperiodic content. The label-probe
response under \(-\)aperiodic is cell-conditional and feeds the
per-cell \(1/f\) role column of Tab.~\ref{tab:verdict_matrix}.

\textbf{The same intervention extends to the fine-tuned
representation.} To test whether FT inherits the frozen
aperiodic dependence, we re-extracted and fine-tuned each FM
end-to-end on both ablated inputs under the same recipe as
the baseline run (3 seeds; Sec.~\ref{sec:methods_anchor}) and
recomputed both probes on the resulting features
(Fig.~\ref{fig:anchor}b, bottom row). On LaBraM and CBraMod,
the FT subject probe drops under \(-\)aperiodic on 7 of 8 (cell,
FM) pairs (mean \(\Delta\) ranging from \(-4\) to \(-20\)\,pp,
with the per-seed range staying below zero). LaBraM \(\times\)
ADFTD is the borderline case (\(-2.0 \pm 4.7\)\,pp). This is
consistent with two properties of that cell: its frozen subject
probe is already near ceiling, and its high recording-level
redundancy gives the encoder multiple non-aperiodic carriers to
amplify. \(-\)Periodic at the FT level shifts the subject probe
by \(|\Delta| \le 2.5\)\,pp on all 8 (cell, LaBraM/CBraMod)
pairs, matching the frozen null and confirming that the FT
carrier is the \(1/f\) component, not narrowband peaks. REVE's
FT subject probe shifts by \(|\Delta| \le 2.5\)\,pp on every
cell under both ablations, preserving the negative control at
the FT representation.

\subsection{Cell-conditional label-side outcomes}
\label{sec:results_regimes}
\label{sec:results_direction}

Two cell-level diagnostics remain. The layer-wise label-probe
trajectory shows how much label-aligned variance survives to the
final pooled feature, and the within-subject direction-consistency
test asks whether subjects share a contrast axis at all. The
combined assessment matrix appears with the Discussion
(Tab.~\ref{tab:verdict_matrix}).

\textbf{Layer-wise label-probe trajectory}
(Sec.~\ref{sec:methods_layerwise}; Fig.~\ref{fig:layerwise}). On
EEGMAT, the label probe is stable across depth
(\(0.69\)--\(0.78\)) for all three FMs. On ADFTD, the label probe
peaks early (relative depth \(\sim 0.17\), BA
\(0.78\)--\(0.83\)) and descends as later blocks compress along
the subject axis without further label gain. On SleepDep, the
label probe is flat near chance (\(0.50\)--\(0.59\)) at every
depth. On Stress, the label probe descends monotonically with
depth on all three FMs (LaBraM \(0.49 \to 0.42\); CBraMod
\(0.49 \to 0.36\); REVE \(0.47 \to 0.41\)): the modest
label-aligned signal in the pre-block embedding does not survive
to the final pooled feature.

\begin{figure*}[!htbp]
\centering
\begin{subfigure}[t]{\linewidth}
  \centering
  \includegraphics[width=\linewidth]{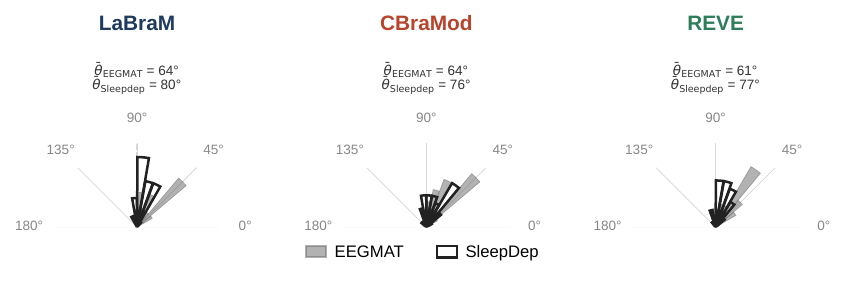}
  \caption{}
  \label{fig:direction:rose}
\end{subfigure}\\[4pt]
\begin{subfigure}[t]{0.49\linewidth}
  \centering
  \includegraphics[width=\linewidth]{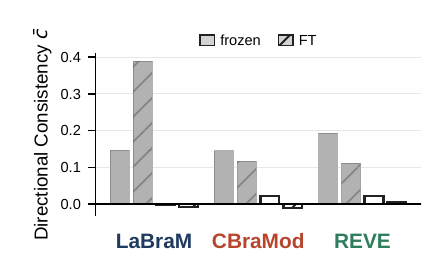}
  \caption{}
  \label{fig:direction:cbar}
\end{subfigure}\hspace{4pt}%
\begin{subfigure}[t]{0.49\linewidth}
  \centering
  \includegraphics[width=\linewidth]{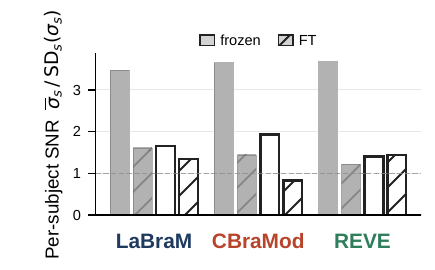}
  \caption{}
  \label{fig:direction:snr}
\end{subfigure}
\caption{\textbf{Within-subject direction and SNR (frozen vs.\ FT).}
For each subject the contrast vector $v_i = \mu_{i,1} - \mu_{i,0}$
is formed in the FM's full feature space; the group consensus is
$v_c = \overline{v_i}/\lVert\overline{v_i}\rVert$. Filled gray =
EEGMAT, outlined black = SleepDep.
\textbf{(\subref{fig:direction:rose})} Polar half-circle rose
(frozen), one panel per FM: $\theta_i = \arccos\langle v_i,
v_c\rangle$; $0^{\circ}$ aligns, $90^{\circ}$ is the
high-dimensional isotropic null.
\textbf{(\subref{fig:direction:cbar})} Group-level $\bar c$
(Eq.~\ref{eq:wsci}) across 3 FMs $\times$ 2 within-subject paired
cells, frozen vs.\ FT. Trait cells (Stress, ADFTD) lack
within-subject contrast by design.
\textbf{(\subref{fig:direction:snr})} Per-subject SNR
(Eq.~\ref{eq:snr_persubj}); dashed line at $\mathrm{SNR}=1$
(signal $\sim$ noise).}
\label{fig:direction}
\end{figure*}

\textbf{Within-subject direction consistency}
(Sec.~\ref{sec:methods_wsci}; Fig.~\ref{fig:direction}). EEGMAT
yields group-level mean pairwise cosine
\(\bar c \in [+0.07, +0.15]\) across three FMs (subjects share a
contrast axis above the isotropic null). SleepDep yields
\(\bar c \approx 0\) (subjects do not share a contrast axis);
cosine PERMANOVA (Sec.~\ref{sec:methods_wsci};
App.~\ref{app:permanova}, Tab.~\ref{tab:permanova_full})
still detects label-associated structure on every (FM, state)
pair (\(p \leq 0.001\)). A local per-subject contrast exists but
does not generalize across subjects (per-cell assessment in
Tab.~\ref{tab:verdict_matrix}). We read these four outcomes as
different sides of a single representation bottleneck
(Sec.~\ref{sec:discussion}).

%% file: tables/main/table2_master_performance.tex
\begin{table*}[htbp]
\centering
\footnotesize
\setlength{\tabcolsep}{4pt}
\caption{\textbf{Subject-disjoint 5-fold CV balanced accuracy across the
four cells.} Rows: classical handcrafted-feature LogReg, four
non-FM deep baselines, and three foundation models under linear
probe (LP) and full fine-tuning (FT). Columns ordered by the
four-cell layout. Values are 3-seed mean with sample std
(seeds $\{42, 123, 2024\}$); \textbf{bold} = global best within
column. Method details: Sec.~\ref{sec:methods_ft}.}
\label{tab:master_performance}
\resizebox{\textwidth}{!}{%
\begin{tabular}{ll cccc}
\toprule
\textbf{Method} & \textbf{Strategy} & \textbf{EEGMAT} & \textbf{ADFTD} & \textbf{SleepDep} & \textbf{Stress} \\
 & & {\scriptsize\textit{within$\times$consensus}} & {\scriptsize\textit{trait$\times$consensus}} & {\scriptsize\textit{within$\times$no-consensus}} & {\scriptsize\textit{trait$\times$no-consensus}} \\
\midrule
Classical LR & -- & \textbf{0.847{\scriptsize\,$\pm$\,0.050}} & 0.587{\scriptsize\,$\pm$\,0.004} & 0.542{\scriptsize\,$\pm$\,0.037} & 0.506{\scriptsize\,$\pm$\,0.019} \\
\midrule
EEGNet & -- & 0.671{\scriptsize\,$\pm$\,0.042} & 0.793{\scriptsize\,$\pm$\,0.016} & 0.500{\scriptsize\,$\pm$\,0.014} & \textbf{0.545{\scriptsize\,$\pm$\,0.059}} \\
ShallowConvNet & -- & 0.699{\scriptsize\,$\pm$\,0.008} & 0.788{\scriptsize\,$\pm$\,0.041} & 0.505{\scriptsize\,$\pm$\,0.040} & 0.476{\scriptsize\,$\pm$\,0.031} \\
DeepConvNet & -- & 0.602{\scriptsize\,$\pm$\,0.021} & 0.773{\scriptsize\,$\pm$\,0.038} & 0.482{\scriptsize\,$\pm$\,0.095} & 0.524{\scriptsize\,$\pm$\,0.066} \\
EEGConformer & -- & 0.676{\scriptsize\,$\pm$\,0.032} & 0.792{\scriptsize\,$\pm$\,0.021} & 0.514{\scriptsize\,$\pm$\,0.048} & 0.530{\scriptsize\,$\pm$\,0.045} \\
\midrule
\multirow{2}{*}{LaBraM} & LP & 0.755{\scriptsize\,$\pm$\,0.070} & \textbf{0.814{\scriptsize\,$\pm$\,0.023}} & 0.542{\scriptsize\,$\pm$\,0.037} & 0.464{\scriptsize\,$\pm$\,0.032} \\
 & FT & 0.718{\scriptsize\,$\pm$\,0.032} & 0.769{\scriptsize\,$\pm$\,0.045} & 0.546{\scriptsize\,$\pm$\,0.045} & 0.429{\scriptsize\,$\pm$\,0.063} \\
\cmidrule{1-6}
\multirow{2}{*}{CBraMod} & LP & 0.708{\scriptsize\,$\pm$\,0.037} & 0.751{\scriptsize\,$\pm$\,0.010} & 0.542{\scriptsize\,$\pm$\,0.050} & 0.440{\scriptsize\,$\pm$\,0.019} \\
 & FT & 0.694{\scriptsize\,$\pm$\,0.014} & 0.772{\scriptsize\,$\pm$\,0.042} & 0.481{\scriptsize\,$\pm$\,0.049} & 0.438{\scriptsize\,$\pm$\,0.024} \\
\cmidrule{1-6}
\multirow{2}{*}{REVE} & LP & 0.755{\scriptsize\,$\pm$\,0.021} & 0.762{\scriptsize\,$\pm$\,0.017} & \textbf{0.556{\scriptsize\,$\pm$\,0.024}} & 0.497{\scriptsize\,$\pm$\,0.019} \\
 & FT & 0.727{\scriptsize\,$\pm$\,0.021} & 0.796{\scriptsize\,$\pm$\,0.009} & 0.542{\scriptsize\,$\pm$\,0.061} & 0.491{\scriptsize\,$\pm$\,0.015} \\
\bottomrule
\end{tabular}%
}
\end{table*}

%% file: sections/discussion.tex
\section{Discussion}
\label{sec:discussion}

The empirical core of this paper is the Identity Trap: across
four small-\(N\) clinical resting-state cohorts, subject identity
dominates the frozen features of all three EEG-FMs we test, and
standard fine-tuning amplifies this subject geometry in 12 of 12
(cell, FM) pairs while amplifying label geometry only in cells
where prior literature has already established a cross-subject EEG
marker. On two of the three FMs, the aperiodic \(1/f\) background
is one identifiable carrier of this subject axis. The four cells map onto four distinct
outcomes (Tab.~\ref{tab:verdict_matrix}), but they share a single
representation bottleneck rather than acting as four unrelated
cases.

\input{tables/main/table3_verdict_matrix.tex}

\subsection{The representation bottleneck}
\label{sec:discussion_bottleneck}

FM features carry stable subject geometry more strongly than
task-aligned variance, and fine-tuning reinforces this ordering
rather than correcting it. The frozen-state evidence has three
parts. The subject-variance fraction is 13--89\(\times\) a
matched random-Gaussian null in 12 of 12 frozen pairs
(Tab.~\ref{tab:null_control}); the layer-wise probe shows
that the subject axis becomes linearly decodable within the
first two to three transformer blocks rather than at the
embedding output, so the encoder actively amplifies subject
identity into a separable direction; and closed-form
subject-axis erasure confirms this dominance sits on a
removable linear axis, collapsing the linear subject probe to
chance in every pair (Tab.~\ref{tab:leace}). On LaBraM and CBraMod, the
aperiodic \(1/f\) background is one identifiable carrier:
removing it drops the linear subject probe by \(9\) to
\(19\)\,pp uniformly across all four cells. REVE shows no measurable aperiodic dependence; the
design-axis differences between REVE and the LaBraM-plus-CBraMod
group are listed in Sec.~\ref{sec:limitations}.

Fine-tuning then amplifies whatever subject axis it inherits.
\(\Delta f_{\mathrm{subj}}\) is positive in every pair (\(+10\)
to \(+63\)\,pp), whereas \(\Delta f_{\mathrm{label}}\) is positive
in 6 of 6 consensus-marker pairs (\(+0.6\) to \(+8.4\)\,pp) and
spans zero in the 6 no-consensus pairs. When no
literature-supported cross-subject marker is available for FT to
amplify toward, FT still adapts, but what it adapts toward is
the subject axis. This connects to a recurring caution in the
EEG-FM literature. Subject-disjoint critiques
\citep{brookshire2024leakage} establish that trial-level splits
inflate accuracy via subject identity. Our diagnostic addresses
the next concern: even under subject-disjoint splitting,
subject-correlated features can still survive the split. The
bottleneck here is not opaque feature noise; it is the stable
presence of a single identifiable component in the input signal. Prior
work identifies subject-disjoint splitting as a necessary
protocol; we show that this protocol alone is not sufficient for
clinical discovery if the model instead aligns its features with
stable subject traits. \textsc{FMScope} provides a representation-level
account of this persistent failure.

\subsection{Spectral-bias amplification under random-head fine-tuning}
\label{sec:discussion_mechanism}

We interpret the cell-conditional amplification asymmetry through
a gradient-geometry reading, now supported on LaBraM and CBraMod
by the FT row of Fig.~\ref{fig:anchor}b. Fine-tuning concentrates supervised
signal along whichever direction already carries the most
variance in the pretrained representation, regardless of
whether that direction aligns with the supervised objective.
Across 12 (cell, FM) pairs, \(\Delta f_{\mathrm{subj}}\) is positive
in every case (\(+10\) to \(+63\)\,pp), and the largest
amplifications fall in the no-consensus trait cell where Stress
reaches \(+32\) to \(+63\)\,pp on the three FMs.
\(\Delta f_{\mathrm{label}}\) is positive in 6 of 6
consensus-marker pairs (\(+0.6\) to \(+8.4\)\,pp) and spans zero
in the 6 no-consensus pairs (\(-1.9\) to \(+0.8\)\,pp). This
asymmetry is what we expect if FT acts on existing axes:
subject-axis amplification is universal because that axis is
large in every frozen representation, whereas label-axis
amplification depends on whether a label-aligned axis is already
large enough to be amplified at all. Under a randomly initialized classification
head, this gradient-geometry reading has formal anchors: linear
probing then fine-tuning analyses show that LP induces a
linear-head-norm regime that preserves the backbone's pretrained
feature directions during the subsequent FT stage
\citep{tomihari2024lpft}, and parameter-efficient adaptation
directions that capture the most activation variance provably
maximize the expected gradient signal \citep{paischer2024eva}. These results combine with classical linear-network results
showing that gradient descent preferentially learns the
largest-singular-value directions of the input--output
correlation \citep{saxe2019linear}. The prescription that follows
from this combination is layer-localized fine-tuning keyed to
shift type \citep{lee2023surgicalft}. Our variance fractions
correspond to the marginalization step of demixed PCA
\citep{kobak2016dpca}, the established systems-neuroscience
predecessor of factor-wise variance accounting.

\subsection{A coding pattern shared across modalities}
\label{sec:discussion_cross_modality}

Our finding sits between two adjacent observations,
neuroscience-side fingerprinting and ML-side shortcut learning,
which converge on the same operational competition between
stable biometric structure and unstable task signal. Our
diagnostic documents the Identity Trap as a specific dissociation:
subject variance dominates the frozen features, label variance is
subordinate, and on LaBraM and CBraMod the aperiodic \(1/f\)
background is one identifiable subject carrier. The same
dissociation between stable subject structure and task signal is
documented across resting-state fMRI fingerprinting and EEG
biometrics (Sec.~\ref{sec:related}); we show it holds inside
EEG-FM representations as well, using variance decomposition as
the test. The broader observation that deep networks
exploit dataset-stable shortcut features over invariant
discriminative cues \citep{geirhos2020shortcut} supplies the
cross-architecture frame. Our diagnostic in turn provides
a physically-grounded instance of that phenomenon in biosignal
foundation models: the shortcut here is not a pure statistical
artifact of the training corpus but has a measurable physiological
component of the input (the aperiodic \(1/f\) background),
high-energy, cross-session stable, and easier to reconstruct
than the clinical signal the model is nominally trained to
support. Concurrent prescriptive work on cross-subject EEG/MEG-FM
adaptation includes SuLoRA, which decomposes each weight matrix
into a shared component and a per-subject low-rank correction
\citep{klein2025sulora}, and SCOPE, a prototype-guided adaptation
framework for label-limited EFM fine-tuning \citep{ma2026scope}.
Both prescribe a fix (adding subject-specific parameters or
prototypes) without measuring the subject axis they adapt to, and
neither relates it to a physiological component of the signal. Our
diagnostic is complementary and upstream: it quantifies the axis in
the frozen feature space, shows it is removable in closed form, and
identifies one physiological carrier. Both adaptation methods are
evaluated on within-subject task contrasts or consensus-marker
datasets; whether they extend to the no-consensus column, which we
predict to be intractable, remains open.

\subsection{Implications and \textsc{FMScope} use}
\label{sec:discussion_clinical}
\label{sec:discussion_framework}
\label{sec:discussion_regimes}

We draw three deployment-side implications from the assessment
matrix. The pre-pilot frozen-feature pass is a cheap filter:
when the assessment falls below linear-probe resolution (Stress
on all three FMs), the bottleneck lies in the recording substrate
itself (an information deficit) rather than in model capacity.
Scaling the pretraining corpus or model size does not address
this class of failure; the remedy lies in how the data are
collected, not in the model.
Trait-cell labels collected within a single session conflate
disease state with stable subject features (aperiodic \(1/f\) being
one identifiable carrier on LaBraM and CBraMod). To separate the
two, the label must change within a subject. Only then does its
discriminative axis fall outside the between-subject subspace that
LEACE erases. Some labels allow this: when the underlying state
changes over time, recording each subject under both conditions
captures the variation. Other labels do not: a dementia diagnosis
is a fixed subject attribute, so no recording protocol can separate
it from identity. For these labels the entanglement is irreducible
at the representation level, and the discriminative features must
instead be validated as physiological rather than biometric.
Scaling model capacity helps in neither case.
Brain--computer-interface calibration claims should pass a
within-subject direction-consistency check before any
subject-independent generalization is asserted; a cell like
SleepDep, where contrasts are idiosyncratic within each subject,
can support high in-subject calibration accuracy with near-zero
cross-subject transfer. These assessments also depend on
reading the diagnostics jointly: the SleepDep assessment is
recoverable only from the disagreement pattern across tools, not
from any single one. Building on the established
subject-disjoint splitting culture in EEG
\citep{brookshire2024leakage,lotte2018}, the assessment matrix
turns this into a frozen-feature pre-pilot test: whether the
features show the geometric signatures (subject dominance,
aperiodic anchoring, axis disagreement) that predict cross-subject
failure before any fine-tuning compute is committed.

\subsection{Limitations and scope}
\label{sec:limitations}

We note five limitations to the present claims. \emph{(i)} Our
\(2{\times}2\) layout relies on one dataset per cell;
consequently, although the \textsc{FMScope} framework is
generalizable, the specific cell-level clinical observations
remain empirical features of these four cohorts and await
broader validation. \emph{(ii)} All four cells are
small by machine-learning standards (Stress \(N=17\) subjects;
EEGMAT and SleepDep \(N=36\); ADFTD \(N=65\)); we mitigate seed
sensitivity with 3-seed FT and 8-seed LP, but per-cell
confidence intervals are wide and pointwise comparisons across
cells are descriptive. The Mann--Whitney U is computed across
12 (cell \(\times\) FM) points drawn from four independent cells;
the column claim rests on the non-overlapping effect-size ranges
(\(+0.6\) to \(+8.4\)\,pp vs \(-1.9\) to \(+0.8\)\,pp), which the
test summarizes.
\emph{(iii)} Our ablations are input-level interventions; they
establish statistical entanglement between aperiodic \(1/f\) and
subject identity, not a causal mechanism. The aperiodic-anchor effect
holds on LaBraM and CBraMod but not on REVE, and these two groups
differ along at least five concurrent design axes that we cannot
factorially separate in an N=3 model panel: (a) the presence of
spectral processing in the pretraining pipeline (LaBraM's
encoder targets discrete codes whose decoder reconstructs Fourier
amplitude+phase; CBraMod's patch embedding adds an FFT-derived branch
to a temporal CNN; REVE operates on raw signal throughout);
(b) reconstruction loss (CBraMod's MSE versus REVE's explicitly
motivated L1, chosen by \citet{reve2025} to avoid L2 amplification of
high-amplitude EEG content); (c) pretraining corpus diversity
(REVE's 25{,}000 subjects across 92 heterogeneous datasets versus
CBraMod's \(\sim\!15{,}000\) subjects from a single clinical source);
(d) positional encoding (REVE's 4D Fourier sinusoidal versus learned
or convolutional alternatives); (e) an attention-pooling secondary
task that REVE adds and the other two omit. We therefore report the
two-versus-one pattern descriptively, not as a mechanism claim; an
architecture-level controlled ablation would be required to isolate
any single axis. \emph{(iv)} All analyses
assume subject-disjoint cross-validation, recording-level
labels, and a \(\geq 19\)-channel 10--20 montage; cells with
finer label resolution or trial-level splitting may produce
different readings. The DASS-21 self-report on Stress
\citep{lovibond1995} is a past-week screener, and we adopt the
per-recording binarization of \citet{komarov2020stress} for
cross-protocol comparability with \citet{wang2025theory}; our
subject-disjoint numbers reproduce that work's protocol-side
configuration but apply each FM's release-default fine-tuning
recipe rather than a per-dataset hyperparameter sweep.
\emph{(v)} Single-session ADFTD cannot separate disease state from
stable subject traits by data alone. The diagnosis is a fixed
subject attribute and does not change within a subject, so
collecting more sessions would not separate them either. For such
labels the limit is intrinsic, and the discriminative features must
be validated against external physiological evidence rather than
addressed with more recordings. We did not pre-align signals using Riemannian or Euclidean
alignment methods before fine-tuning. This engineering step may
attenuate the subject-axis amplification and is a natural
follow-up, but it is not a prerequisite for the diagnostic claim
itself.

\subsection{Future directions}
\label{sec:discussion_future}

We see three algorithmic directions following from the
bottleneck reading: FOOOF-detrended pretraining (the causal test
of whether aperiodic \(1/f\) is the masked objective's primary
subject carrier), explicit gradient-reversal objectives at
fine-tuning \citep{ganin2015} that could augment existing
general-purpose SSL pretraining recipes for EEG
\citep{banville2021ssl,kostas2021bendr}, and explicit
cross-subject alignment regularization at pretraining. On the
empirical side, replication with a second dataset per cell, a
layer-wise FOOOF-aperiodic probe, a broader
FM panel that varies the five design axes independently, and a
within-subject paired protocol on a no-consensus label would
sharpen the present descriptive observations into axis-level
claims. Integrating \textsc{FMScope}-style diagnostics into
large-scale community benchmarks
\citep{aristimunha2506eeg,eegbench2025,brain4fms2026} would
complement leaderboard accuracy with a representation-level
check that distinguishes models rewarded for learning clinical
biomarkers from those rewarded for optimizing subject-specific
aperiodic anchors. The diagnostic framework we report here
characterizes the substrate; pretraining-objective work is the
path that moves the substrate.

%% file: tables/main/table3_verdict_matrix.tex
% Tab 3 — Cross-diagnostic results summary (numeric, mean across 3 FMs)
\begin{table*}[htbp]
\centering
\small
\caption{\textbf{Cross-diagnostic results and per-cell assessment.}
Numeric summary of \textsc{FMScope}'s four diagnostics per cell.
All values are the arithmetic mean across the three foundation models
(LaBraM, CBraMod, REVE); per-FM values appear in
Tab.~\ref{tab:variance_full} (\(\Delta f_{\mathrm{label}}\)),
Fig.~\ref{fig:layerwise} (layer probe), Fig.~\ref{fig:direction}
(\(\bar c\)), and Tab.~\ref{tab:fooof_full} (\(1/f\) ablation drops).
Columns: \(\Delta f_{\mathrm{label}}\) = FT minus frozen
label-explained variance fraction; \emph{layer probe BA max\,/\,last}
= max and last-layer balanced accuracy of the layer-wise label probe;
\(\bar c\) = median within-subject direction-consistency (``---'' in
trait cells where no within-subject paired contrast exists);
\emph{\(1/f\) drops state\,/\,subj} = drops in label-state and
subject-identity probe BAs after aperiodic-component ablation
(positive values mean the ablated component carried that signal).
\emph{Outcome} is the qualitative cell-level reading
(Sec.~\ref{sec:results_regimes}); the four outcomes are exhaustive
over the 2$\times$2 cell layout: W = within-subject paired,
T = subject-label trait, C = consensus cross-subject marker,
N = no consensus marker.}
\label{tab:verdict_matrix}
\renewcommand{\arraystretch}{1.2}
\setlength{\tabcolsep}{4pt}
\begin{tabular}{@{}lcccccl@{}}
\toprule
\multirow{2}{*}{\textbf{Cell}} &
\multirow{2}{*}{\textbf{$\Delta f_{\mathrm{label}}$}} &
\multicolumn{2}{c}{\textbf{layer probe BA}} &
\multirow{2}{*}{\textbf{$\bar c$}} &
\textbf{$1/f$ drops} &
\multirow{2}{*}{\textbf{Outcome}} \\
\cmidrule(lr){3-4}
 & & \textbf{max} & \textbf{last} & & \textbf{state\,/\,subj} & \\
\midrule
EEGMAT \textit{(W,\,C)}    & $+0.043$ & $0.77$ & $0.74$ & $+0.16$ & $+0.04 / +0.13$ & \textbf{Cross-subject-aligned}        \\
ADFTD \textit{(T,\,C)}     & $+0.041$ & $0.82$ & $0.78$ & ---     & $+0.00 / +0.07$ & \textbf{Label--subject coupled}       \\
SleepDep \textit{(W,\,N)}  & $-0.008$ & $0.58$ & $0.55$ & $+0.01$ & $-0.04 / +0.12$ & \textbf{Idiosyncratic within-subject} \\
Stress \textit{(T,\,N)}    & $+0.003$ & $0.50$ & $0.40$ & ---     & $-0.01 / +0.06$ & \textbf{Below linear-probe resolution}\\
\bottomrule
\end{tabular}
\end{table*}

%% file: sections/conclusion.tex
\section{Conclusion}
\label{sec:conclusion}

Subject-disjoint cross-validation is necessary but not sufficient
for representation discovery in small-\(N\) clinical resting-state
EEG. Across the 12 (cell, FM) pairs we examined, the frozen
embedding already placed subject identity on its largest-variance
direction, and fine-tuning amplified that direction whether or not
the label aligned with it; a high accuracy is therefore consistent
either with a transferable clinical biomarker or with subject
identity that merely co-varies with the label in this cohort, and
accuracy alone cannot separate the two. \textsc{FMScope} is a
frozen-feature pre-flight check, complementary to subject-disjoint
splitting, that tells the user before any fine-tuning whether a
cell's representation carries label-aligned biological structure
or subject identity dressed up as a label. Where it signals an
information deficit rather than a capacity limit, the remedy is
methodological: a within-subject label should be recorded under
both conditions so that its discriminative axis separates from
between-subject identity, whereas a label fixed at the subject
level cannot be disentangled by any protocol, and its
discriminative features must instead be shown to correspond to an
independently established physiological marker rather than to
incidental subject traits. The framework generalizes; the per-cell
outcomes are empirical features of these four cohorts and three
FMs, and broader validation awaits future cohorts.

%% file: sections/statements.tex
\section*{Data availability}
This study uses four previously collected EEG datasets. EEGMAT
\citep{eegmat} is available from PhysioNet
(\url{https://physionet.org/content/eegmat/1.0.0/}); the ADFTD
dataset \citep{adftd} is available from OpenNeuro (ds004504); the
sleep-deprivation rsEEG dataset \citep{xiang2025sleepdep} is
available as published by the original authors. The resting-state
stress dataset \citep{komarov2020stress} is not currently
deposited in a public repository; it was shared by the originating
laboratory (co-author T.-P. Jung's group) for the present
secondary analysis and is available from the authors upon
reasonable request and with permission from the original
investigators. No new data were collected.

\section*{Code availability}
The \textsc{FMScope} toolkit, comprising the five frozen-representation
diagnostics used in this paper (variance decomposition, subject-axis
erasure, aperiodic FOOOF ablation, layer-wise probe, and within-subject
direction consistency), together with the exact scripts that reproduce
every table and figure, is available at
\url{https://github.com/Jimmy110101013/fmscope}. Reproduction runs from
bundled aggregate results and frozen-feature caches, without raw EEG or
model weights. Pretrained foundation-model weights (LaBraM, CBraMod,
REVE) are obtained from the respective original repositories and are not
redistributed by this work.

\section*{Ethics statement}
This work uses previously collected, de-identified EEG datasets.
EEGMAT, ADFTD, and the sleep-deprivation dataset are publicly
released; the Komarov et al.\ stress dataset was obtained from
the original investigators (co-author T.-P. Jung's lab) for
secondary analysis. The original data-collection studies were
conducted under their respective IRB approvals and consent
procedures, which we cite. No new human-subject data were
collected, and no additional IRB approval was required.

\section*{Conflict of interest}
The authors declare no competing interests.

\section*{Funding}
This work received no external funding.

\section*{Author contributions}
J.-Y.L.: conceptualization, methodology, software, formal
analysis, investigation, data curation, writing --- original
draft, writing --- review and editing, visualization. T.-P.J.:
supervision, conceptual input, writing --- review and editing.

%% file: sections/appendix.tex
\appendix
\section*{Appendix}
\renewcommand{\thesection}{\Alph{section}}
\setcounter{section}{0}
\renewcommand{\thetable}{A\arabic{table}}
\setcounter{table}{0}

\section{Variance decomposition: per-pair values}
\label{app:variance_full}

Window-level variance decomposition, full per-(cell, FM) values for
the 12 frozen and 12 fine-tuned pairs underlying
Sec.~\ref{sec:results_subject_dominant} and
Fig.~\ref{fig:variance_2x2}.
\(\Delta\) columns are FT minus frozen.

\begin{table}[!htbp]
\centering
\small
\setlength{\tabcolsep}{4pt}
\caption{Per-pair label-fraction and subject-fraction values
(window-level variance decomposition). Frozen and FT in percentage
points; \(\Delta\) = FT \(-\) frozen.}
\label{tab:variance_full}
\begin{tabular}{llrrrrrr}
\toprule
\multirow{2}{*}{\textbf{Cell}} & \multirow{2}{*}{\textbf{FM}} &
\multicolumn{2}{c}{\textbf{Frozen \%}} &
\multicolumn{2}{c}{\textbf{FT \%}} &
\multicolumn{2}{c}{\textbf{\(\Delta\) (pp)}} \\
\cmidrule(lr){3-4} \cmidrule(lr){5-6} \cmidrule(lr){7-8}
& & label & subj & label & subj & label & subj \\
\midrule
\multirow{3}{*}{EEGMAT}    & LaBraM  & 1.3 & 26.8 &  9.7 & 45.4 & \(+8.4\) & \(+18.6\) \\
                            & CBraMod & 1.1 & 29.6 &  2.6 & 60.3 & \(+1.5\) & \(+30.7\) \\
                            & REVE    & 1.8 & 29.7 &  4.6 & 50.0 & \(+2.9\) & \(+20.3\) \\
\midrule
\multirow{3}{*}{ADFTD}     & LaBraM  & 6.6 & 55.4 & 13.3 & 78.9 & \(+6.6\) & \(+23.5\) \\
                            & CBraMod & 5.5 & 58.8 & 10.7 & 81.6 & \(+5.2\) & \(+22.8\) \\
                            & REVE    & 4.6 & 46.5 &  5.2 & 80.5 & \(+0.6\) & \(+34.0\) \\
\midrule
\multirow{3}{*}{SleepDep}  & LaBraM  & 0.5 & 30.8 &  0.3 & 51.3 & \(-0.1\) & \(+20.4\) \\
                            & CBraMod & 0.5 & 39.3 &  0.3 & 49.4 & \(-0.3\) & \(+10.1\) \\
                            & REVE    & 2.2 & 32.4 &  0.3 & 51.2 & \(-1.9\) & \(+18.7\) \\
\midrule
\multirow{3}{*}{Stress}    & LaBraM  & 2.9 & 20.7 &  2.5 & 52.8 & \(-0.5\) & \(+32.0\) \\
                            & CBraMod & 1.4 & 18.5 &  2.2 & 81.2 & \(+0.8\) & \(+62.7\) \\
                            & REVE    & 2.1 & 20.7 &  2.6 & 64.0 & \(+0.5\) & \(+43.3\) \\
\bottomrule
\end{tabular}
\end{table}

The \(\Delta f_{\mathrm{label}}\) column-effect Mann--Whitney
(method: Sec.~\ref{sec:methods_variance}) operates on the 12
\(\Delta\)label values above. Cosine PERMANOVA per-cell
assessments are reported in App.~\ref{app:permanova}.

\section{Random-Gaussian null control for the Identity-Trap inequality}
\label{app:null_control}

The raw inequality \(f_{\mathrm{subj}} > f_{\mathrm{label}}\) on
a crossed sum-of-squares decomposition is combinatorially
guaranteed whenever the number of subjects \(S\) exceeds the
number of labels \(L\): for an iid Gaussian embedding with \(N\)
rows the marginal sum-of-squares fractions converge to their
degrees of freedom,
\(\mathrm{E}[f_{\mathrm{subj}}] \approx (S{-}1)/(N{-}1)\) and
\(\mathrm{E}[f_{\mathrm{label}}] \approx (L{-}1)/(N{-}1)\), with
\(S > L\) for every cell in our panel. We therefore quantify the
Identity-Trap evidence as the \emph{excess of the frozen
\(f_{\mathrm{subj}}\) over a matched random-Gaussian null} of
identical shape \((N, D)\), averaged over \(K=20\) seeds.
Tab.~\ref{tab:null_control} reports per-pair excess ratios; the
empirical null mean agrees with the closed-form
\((S{-}1)/(N{-}1)\) prediction to three decimal places.

\begin{table}[!htbp]
\centering
\small
\setlength{\tabcolsep}{4pt}
\caption{Frozen subject-variance fraction (real FM vs.\ matched
random-Gaussian null of identical shape) for the 12 (cell, FM)
pairs. \(N\) is the number of windows, \(S\) the number of
subjects, \(L=2\) for every cell. ``Null \(f_{\mathrm{subj}}\)''
is the mean \(\pm\) SD over \(K=20\) iid Gaussian draws; ``df pred''
is the closed-form \((S{-}1)/(N{-}1)\); ``Excess \(\times\)'' is
real / null mean. \(\Delta f_{\mathrm{label}}\) (Tab.~\ref{tab:variance_full})
is intrinsically null-corrected because the combinatorial offset
cancels in the frozen-to-FT difference.}
\label{tab:null_control}
\begin{tabular}{llrrrrrr}
\toprule
\textbf{Cell} & \textbf{FM} & \(N\) & \(S\) & \textbf{Real \(f_{\mathrm{subj}}\) (\%)} & \textbf{Null \(f_{\mathrm{subj}}\) (\%)} & \textbf{df pred (\%)} & \textbf{Excess (\(\times\))} \\
\midrule
\multirow{3}{*}{EEGMAT}    & LaBraM  & 1707 & 36 & 26.85 & 2.04\,$\pm$\,0.02 & 2.05 & 13.2 \\
                            & CBraMod & 1707 & 36 & 29.56 & 2.06\,$\pm$\,0.03 & 2.05 & 14.3 \\
                            & REVE    & 1707 & 36 & 29.74 & 2.05\,$\pm$\,0.03 & 2.05 & 14.5 \\
\midrule
\multirow{3}{*}{ADFTD}     & LaBraM  & 9701 & 65 & 55.40 & 0.66\,$\pm$\,0.01 & 0.66 & 83.9 \\
                            & CBraMod & 9701 & 65 & 58.76 & 0.66\,$\pm$\,0.01 & 0.66 & 89.3 \\
                            & REVE    & 9701 & 65 & 46.49 & 0.66\,$\pm$\,0.01 & 0.66 & 70.4 \\
\midrule
\multirow{3}{*}{SleepDep}  & LaBraM  & 4207 & 36 & 30.83 & 0.84\,$\pm$\,0.01 & 0.83 & 36.7 \\
                            & CBraMod & 4207 & 36 & 39.27 & 0.83\,$\pm$\,0.02 & 0.83 & 47.3 \\
                            & REVE    & 4207 & 36 & 32.45 & 0.83\,$\pm$\,0.01 & 0.83 & 39.1 \\
\midrule
\multirow{3}{*}{Stress}    & LaBraM  & 4421 & 14 & 20.75 & 0.29\,$\pm$\,0.01 & 0.29 & 70.7 \\
                            & CBraMod & 4421 & 14 & 18.46 & 0.30\,$\pm$\,0.01 & 0.29 & 62.6 \\
                            & REVE    & 4421 & 14 & 20.69 & 0.29\,$\pm$\,0.00 & 0.29 & 70.1 \\
\bottomrule
\end{tabular}
\end{table}

\section{Cosine PERMANOVA: per-cell label-structure detection}
\label{app:permanova}

Per-cell PERMANOVA \(p\)-values on the variance-triangulation
cache (method: Sec.~\ref{sec:methods_wsci}).

\begin{table}[!htbp]
\centering
\small
\setlength{\tabcolsep}{6pt}
\caption{Cosine PERMANOVA \(p_{\text{label}}\) per cell \(\times\)
FM \(\times\) state. Floor \(\approx 0.001\) (\(999\)
permutations). EEGMAT, ADFTD, and SleepDep clear the floor on
every (FM, state) cell; Stress is null on every cell.}
\label{tab:permanova_full}
\begin{tabular}{llcc}
\toprule
\textbf{Cell} & \textbf{FM} & \textbf{Frozen} \(p\) & \textbf{FT} \(p\) \\
\midrule
\multirow{3}{*}{EEGMAT}    & LaBraM  & 0.001 & 0.001 \\
                            & CBraMod & 0.001 & 0.001 \\
                            & REVE    & 0.001 & 0.001 \\
\midrule
\multirow{3}{*}{ADFTD}     & LaBraM  & 0.001 & 0.001 \\
                            & CBraMod & 0.004 & 0.001 \\
                            & REVE    & 0.001 & 0.002 \\
\midrule
\multirow{3}{*}{SleepDep}  & LaBraM  & 0.001 & 0.001 \\
                            & CBraMod & 0.001 & 0.001 \\
                            & REVE    & 0.001 & 0.001 \\
\midrule
\multirow{3}{*}{Stress}    & LaBraM  & 0.165 & 0.647 \\
                            & CBraMod & 0.477 & 0.957 \\
                            & REVE    & 0.417 & 0.933 \\
\bottomrule
\end{tabular}
\end{table}

PERMANOVA on SleepDep is significant on every (FM, state) pair
even though \(\Delta f_{\mathrm{label}}\) is negative across all
three FMs. This is consistent with the within-subject paired
design: each pair block is anchored to one subject, and the
per-pair contrast leaves a label-detectable signature in the
cosine geometry whether or not fine-tuning amplifies it. We
therefore use PERMANOVA as a within-cell detector of
label-related structure rather than as the axis-level test; the
consensus-vs-no-consensus contrast is given by the
\(\Delta f_{\mathrm{label}}\) Mann--Whitney U above. On Stress,
PERMANOVA shows no effect on any (FM, state) pair, which supports
the below-linear-probe-resolution assessment
(Tab.~\ref{tab:verdict_matrix}) without relying on the layer-wise
probe alone.

\section{FOOOF aperiodic / periodic ablation: full probe BA table}
\label{app:fooof_full}

Label and subject probe baseline BAs and ablation deltas for the
4 cells \(\times\) 3 FMs underlying
Sec.~\ref{sec:results_aperiodic} and Fig.~\ref{fig:anchor}. The
subject probe is a 5-fold temporal-block LDA with Ledoit--Wolf
shrinkage. All values in BA \(\times 100\); \(\Delta\) values in
percentage points.

\begin{table}[!htbp]
\centering
\small
\setlength{\tabcolsep}{4pt}
\caption{Label probe and subject probe baseline BA, with deltas
under FOOOF \(-\)aperiodic and \(-\)periodic. Label probe:
subject-disjoint 5-fold logistic regression on per-window features.
Subject probe: 5-fold temporal-block LDA. \(\Delta\) values are
ablation BA minus baseline BA.}
\label{tab:fooof_full}
\begin{tabular}{llrrrrrr}
\toprule
\multirow{2}{*}{\textbf{Cell}} & \multirow{2}{*}{\textbf{FM}} &
\multicolumn{3}{c}{\textbf{Label probe}} &
\multicolumn{3}{c}{\textbf{Subject probe}} \\
\cmidrule(lr){3-5} \cmidrule(lr){6-8}
& & base & \(\Delta\)\,aper. & \(\Delta\)\,per. & base & \(\Delta\)\,aper. & \(\Delta\)\,per. \\
\midrule
\multirow{3}{*}{EEGMAT}    & LaBraM  & 76.2 & \(-8.3\) & \(-0.4\) & 87.6 & \(-18.4\) & \(+0.1\) \\
                            & CBraMod & 70.7 & \(-1.0\) & \(+0.2\) & 92.6 & \(-18.9\) & \(-0.1\) \\
                            & REVE    & 75.7 & \(-2.8\) & \(-0.0\) & 99.0 & \(-0.2\)  & \(+0.0\) \\
\midrule
\multirow{3}{*}{ADFTD}     & LaBraM  & 80.7 & \(-2.8\) & \(+0.3\) & 88.3 & \(-10.9\) & \(-0.0\) \\
                            & CBraMod & 74.7 & \(+3.1\) & \(+0.5\) & 92.8 & \(-10.2\) & \(+0.2\) \\
                            & REVE    & 79.5 & \(-0.1\) & \(-0.1\) & 99.3 & \(+0.1\)  & \(+0.0\) \\
\midrule
\multirow{3}{*}{SleepDep}  & LaBraM  & 55.7 & \(+2.8\) & \(-0.3\) & 78.9 & \(-17.9\) & \(-0.1\) \\
                            & CBraMod & 54.9 & \(+6.9\) & \(+0.0\) & 86.1 & \(-17.1\) & \(-0.4\) \\
                            & REVE    & 55.4 & \(+2.1\) & \(+0.1\) & 95.6 & \(-1.2\)  & \(+0.0\) \\
\midrule
\multirow{3}{*}{Stress}    & LaBraM  & 42.3 & \(+0.6\) & \(-0.2\) & 67.3 & \(-9.1\)  & \(-0.0\) \\
                            & CBraMod & 37.7 & \(+5.1\) & \(+0.4\) & 73.5 & \(-9.3\)  & \(+0.0\) \\
                            & REVE    & 39.4 & \(-1.9\) & \(+0.1\) & 92.6 & \(-0.0\)  & \(+0.1\) \\
\bottomrule
\end{tabular}
\end{table}

The \(-\)aperiodic intervention drops the subject probe by
\(9\)--\(19\)\,pp on LaBraM and CBraMod uniformly across all four
cells, while leaving REVE's already-saturated subject probe
(baseline \(\geq 0.93\)) unchanged. Periodic peak removal shifts
either probe by \(\leq 0.8\)\,pp on every (FM, cell) pair.
Cell-conditional label-probe response is interpreted in
Sec.~\ref{sec:results_aperiodic}.

\section{Subject-axis erasure across independent cohorts}
\label{app:erasure_gen}

The single EEGMAT erasure demonstration generalises. Applying the
identical procedure (freeze, LEACE-erase the subject subspace, label
BA pre/post under subject-level CV) to four further marker families
and the five external audit cohorts (Tab.~\ref{tab:erasure_gen}),
\(\Delta_{\mathrm{erase}}\) is positive on all three FMs wherever a
literature-established consensus marker exists --- arithmetic
\(\theta\), motor-imagery ERD, auditory P300 --- and
mixed-to-negative for the two external cohorts whose markers are not
established (SAM40, TDBRAIN-state); the three subject-trait cohorts
have \(\Delta_{\mathrm{erase}}\) undefined but show the same subject
probe collapse to chance. Magnitudes are not comparable across rows
(windows \(1\)--\(5\)\,s, ceiling effects, recording- vs window-level
scoring), but the signs are. Across the twelve consensus-marker
cells (four cohorts \(\times\) three FMs, the primary EEGMAT cell
included) \(\Delta_{\mathrm{erase}}\) is positive in all twelve
(one-sided sign test \(p = 2.4\times10^{-4}\); consensus vs.\
non-consensus cohorts, one-sided Mann--Whitney \(U\),
\(p = 2.2\times10^{-4}\)). Erasing identity thus aids the label
exactly where a cross-subject marker exists, reinforcing
Sec.~\ref{sec:results_subject_dominant} (Tab.~\ref{tab:leace}).

\begin{table}[!htbp]
\centering
\small
\setlength{\tabcolsep}{4pt}
\caption{Subject-axis erasure beyond the EEGMAT demonstration
(Tab.~\ref{tab:leace}). Raw BA: pre-erasure label BA \(\times 100\)
(range over the three FMs). \(\Delta_{\mathrm{erase}}\): post\,\(-\)\,pre
change (pp, 3-seed mean). Top: established consensus markers; bottom:
external cohorts with non-established markers.}
\label{tab:erasure_gen}
\begin{tabular}{lcccc}
\toprule
\multirow{2}{*}{\textbf{Cohort}} & \multirow{2}{*}{\textbf{Raw BA}} &
\multicolumn{3}{c}{\(\Delta_{\mathrm{erase}}\) (pp)} \\
\cmidrule(lr){3-5}
& & LaBraM & CBraMod & REVE \\
\midrule
EEGMAT \citep{eegmat}              & 71--76 & \(+12.0\) & \(+6.0\)  & \(+7.4\)  \\
Test--retest \citep{wang2022testretest} & 68--78 & \(+4.2\)  & \(+4.4\)  & \(+13.9\) \\
EEGMMIDB \citep{schalk2004bci2000} & 61--73 & \(+17.5\) & \(+8.3\)  & \(+26.7\) \\
Aud.--Vis.\ Shift \citep{ceponiene2008} & 74--86 & \(+24.2\) & \(+21.6\) & \(+14.4\) \\
\midrule
SAM40 \citep{ghosh2022sam40}       & 61--69 & \(-5.2\)  & \(-7.3\)  & \(+4.6\)  \\
TDBRAIN-state \citep{vandijk2022tdbrain} & 53--68 & \(-7.9\)  & \(-18.4\) & \(-0.4\)  \\
\bottomrule
\end{tabular}
\end{table}